\UseRawInputEncoding
\pdfoutput=1

\documentclass[10pt,journal,compsoc]{IEEEtran}

\IEEEoverridecommandlockouts
\usepackage{caption}
\usepackage{subcaption}
\usepackage{cite}
\usepackage{amsmath,amssymb,amsfonts}
\usepackage{algorithmic}
\usepackage{graphicx}
\usepackage{textcomp}
\usepackage{xcolor}
\usepackage{booktabs}
\usepackage{multirow} 
\def\BibTeX{{\rm B\kern-.05em{\sc i\kern-.025em b}\kern-.08em
    T\kern-.1667em\lower.7ex\hbox{E}\kern-.125emX}}

\DeclareMathOperator*{\argmax}{arg\,max}

\usepackage{algorithm}

\newtheorem{theorem}{Theorem}
\newtheorem{lemma}{Lemma}
\newtheorem{assumption}{Assumption}

\newtheorem{remark}{Remark}

\long\def\comment#1{}

\newfont{\bbb}{msbm10 scaled 700}

\newfont{\bb}{msbm10 scaled 1100}

\newcommand{\EE}{\mbox{\bb E}}

\newcommand{\qv}{{\bf q}}

\newcommand{\Ac}{{\cal A}}

\newcommand{\Dc}{{\cal D}}

\newcommand{\Ic}{{\cal I}}

\newcommand{\Lc}{{\cal L}}
\newcommand{\Mc}{{\cal M}}
\newcommand{\Nc}{{\cal N}}

\newcommand{\Pc}{{\cal P}}

\newcommand{\Sc}{{\cal S}}
\newcommand{\Tc}{{\cal T}}

\begin{document}

\title{Federated Reinforcement Learning in Heterogeneous Environments\\
}

\author{Ukjo~Hwang,~\IEEEmembership{Student Member,~IEEE} and Songnam~Hong,~\IEEEmembership{Member,~IEEE} 
\IEEEcompsocitemizethanks{\IEEEcompsocthanksitem U. Hwang and S. Hong are with the Department of Electronic Engineering, Hanyang University, Seoul, 04763, Korea.\protect\\
E-mail: \{yd1001,snhong\}@hanyang.ac.kr



}
}




\IEEEtitleabstractindextext{%
\begin{abstract}
We investigate a Federated Reinforcement Learning with Environment Heterogeneity (FRL-EH) framework, where local environments exhibit statistical heterogeneity. Within this framework, agents collaboratively learn a global policy by aggregating their collective experiences while preserving the privacy of their local trajectories. To better reflect real-world scenarios, we introduce a {\em robust} FRL-EH framework by presenting a novel global objective function. This function is specifically designed to optimize a global policy that ensures robust performance across heterogeneous local environments and their plausible perturbations. We propose a tabular FRL algorithm named FedRQ and theoretically prove its asymptotic convergence to an optimal policy for the global objective function. Furthermore, we extend FedRQ to environments with continuous state space through the use of expectile loss, addressing the key challenge of minimizing a value function over a continuous subset of the state space. This advancement facilitates the seamless integration of the principles of FedRQ with various Deep Neural Network (DNN)-based RL algorithms. Extensive empirical evaluations validate the effectiveness and robustness of our FRL algorithms across diverse heterogeneous environments, consistently achieving superior performance over the existing state-of-the-art FRL algorithms. 
\end{abstract}

\begin{IEEEkeywords}
Federated reinforcement learning, robust reinforcement learning, heterogeneous environments.
\end{IEEEkeywords}}

\maketitle

\IEEEdisplaynontitleabstractindextext

\IEEEpeerreviewmaketitle

\IEEEraisesectionheading{\section{Introduction}\label{sec:introduction}}

Reinforcement Learning (RL) has demonstrated remarkable efficacy in tackling complex challenges across various domains, including gaming, robotics, intelligent networks, manufacturing, and finance \cite{gosavi2009reinforcement, arulkumaran2017deep, luong2019applications}. However, the practical implementation of RL algorithms often encounters persistent obstacles, particularly the scarcity of training samples, especially in large action and state spaces. In these situations, acquiring a sufficient number of samples for effective learning becomes increasingly difficult, as thoroughly exploring all potential states is nearly impossible. Therefore, promoting collaboration among multiple RL agents engaged in analogous tasks is crucial for significantly improving learning accuracy and accelerating convergence.

These challenges can be effectively addressed through the deployment of Federated Reinforcement Learning (FRL) \cite{nadiger2019federated, khodadadian2022federated, qi2021federated, xie2023fedkl, lim2021federated, jin2022federated}. The objective of FRL, which integrates RL with Federated Learning (FL) \cite{hong2021communication, gogineni2022communication, kwon2023tighter, hong2023online, salgia2024communication}, is to learn a global policy for sequential-decision making problems through the collaboration of multiple agents. These agents interact with their respective local environments and transmit local updates, such as the gradients of local models, to a central server. This process is designed to ensure that their collected experiences, which may contain sensitive information, remain undisclosed, thereby preserving privacy. A foundational FRL framework was introduced in \cite{nadiger2019federated}, which encompasses the grouping policy, learning policy, and federation policy. In this context, RL is employed to demonstrate the feasibility of fine-grained personalization, while FL is utilized to improve sample efficiency. The effectiveness of federated tabular Q-learning was thoroughly investigated in \cite{khodadadian2022federated}. Additionally, various strategies have been proposed to enhance performance and manage data heterogeneity, including regularization techniques based on Kullback-Leibler divergence \cite{xie2023fedkl} and novel combining weights for FRL \cite{lim2021federated}.

Most prior works assume that local environments are statistically identical, sharing a common state transition probability distribution.  However, in real-world applications, agents often operate in diverse environments. For example, when training a global policy for drone control within the FRL framework, distributed agents collect trajectories or training samples under varying weather conditions. This leads to heterogeneous environments characterized by subtly distinct state transition dynamics. Furthermore, a trained agent is expected to generalize across diverse testing environments with different state transition dynamics; for instance, a drone must operate effectively under changing conditions, such as fluctuations in wind speed and direction. Motivated by this challenge, we in this paper aim to develop a global policy that achieves stable performance across heterogeneous local environments, as well as in potentially perturbed environments.

\subsection{Related Works}\label{subsec:RelatedWorks}

Traditionally, RL focuses on maximizing the cumulative reward within an environment characterized by a fixed state transition dynamic~\cite{sutton1988learning, watkins1992q}. Among various approaches to RL, model-free (or sample-based) RL algorithms are widely adopted, as they do not require explicit knowledge of the environment's dynamics, such as state transition probabilities or reward functions. This characteristic makes them particularly well-suited for complex environments where accurate modeling is challenging.  One of the most prominent model-free algorithms for discrete action spaces is Q-learning~\cite{watkins1989learning}, which was subsequently extended to the Deep Q-Network (DQN)~\cite{mnih2013playing} by incorporating Deep Neural Networks (DNNs) to approximate the Q function. For continuous action spaces, the policy gradient method was introduced in~\cite{sutton1999policy}, and later extended to deterministic policies via the Deterministic Policy Gradient (DPG) algorithm~\cite{silver2014deterministic}. This approach was further developed into the Deep Deterministic Policy Gradient (DDPG) \cite{lillicrap2015continuous}, which employs DNNs to approximate both the policy function and the Q function.

The primary objective of FL is to collaboratively train a global model across multiple clients while rigorously preserving data privacy and security~\cite{li2021survey, li2020federatedlearning, qi2021federated, yang2019federated}. During the training process, only model-related information, such as parameters of a Deep Neural Network (DNN), is exchanged among participating clients, ensuring that raw data remains strictly local and confidential. In the model aggregation phase, the contributions from diverse local datasets enhance the generalization capability of the global model. To reduce communication overhead, most FL algorithms allow clients to perform multiple local updates before periodically synchronizing with a central server via model aggregation~\cite{mcmahan2017communication, li2020federatedoptimization}. The de facto algorithm, referred to as FedAvg, achieves competitive learning performance while incurring lower communication overhead. One of the central challenges in the development of FL is the presence of {\em data heterogeneity} across clients \cite{mendieta2022local, dai2023tackling}. Unfortunately, FedAvg suffers from `client-drift' when data is heterogeneous, resulting in unstable and slow convergence. To effectively address data heterogeneity, several FL algorithms have been proposed, including FedProx \cite{li2020federated}, SCAFFOLD \cite{karimireddy2020scaffold}, FedNova \cite{wang2021novel}, which are typically constructed by appropriately enhancing local update mechanism of FedAvg.

FRL refers to the integration of FL principles into RL settings.  In this framework, each agent interacts solely with its own environment, which is independent from those of other agents, while sharing a common state and action space to collaboratively address analogous tasks. The actions taken by each agent only affect its own environment and yield corresponding rewards. As it is often infeasible for a single agent to explore the entire state space of its environment, collaboration among multiple agents can expedite training and enhance model performance through shared experiences. Similar to FL, FRL strictly prohibits the exchange of row local data (or trajectories) to preserve privacy. Instead, communication is limited to model-related information, such as model parameters or gradients, which is exchanged among agents through a central server~\cite{zhuo2019federated, qi2021federated, liu2019lifelong, nadiger2019federated, wang2020federated}.

Building upon the standard FRL framework, FRL with Environment Heterogeneity (FRL-EH) has been introduced to more accurately reflect the characteristics of real-world scenarios~\cite{jin2022federated}. Whereas standard FRL typically assumes that local environments are statistically identical, FRL-EH relaxes this assumption by allowing each agent to operate within its own distinct environment, thereby accommodating statistical heterogeneity across local environments. This relaxation makes FRL-EH a more realistic and practically applicable framework. The primary objective of FRL-EH is to learn a global policy that ensures robust performance across all heterogeneous local environments while preserving data privacy. Within the FRL-EH framework, QAvg was proposed in~\cite{jin2022federated} by replacing the local update mechanism of FL with the standard Q-learning update rule. QAvg serves as the de facto algorithm for FRL-EH, playing a role analogous to that of FedAvg in conventional FL. It seeks to address environment heterogeneity solely through the global averaging of locally updated Q functions. However, as the degree of heterogeneity increases, this naive averaging approach may result in performance degradation---an issue that mirrors the challenges observed in FedAvg under data heterogeneity.

\subsection{Contribution}\label{subsec:Contribution}

We investigate the phenomenon of {\em environment heterogeneity} within the FRL-EH framework. QAvg, the de facto FRL algorithm in this setting, tackles environment heterogeneity solely through global averaging of locally updated Q functions, directly applying the standard Q-learning update during the local training phase. Given that environment heterogeneity is analogous to the well-established challenge of data heterogeneity in traditional FL, it is natural to hypothesize that QAvg could similarly benefit from more sophisticated local update mechanisms.

Motivated by this insight, we propose scalable and practical FRL algorithms that incorporate robust local update strategies. These algorithms are designed to learn a global policy that achieves consistent performance across diverse environments, including those not encountered during training. As a result, the learned policy is particularly well-suited for safety-critical FRL applications, such as autonomous driving and robotics~\cite{gu2024review}. Our main contributions are outlined as follows:
\begin{itemize}
    \item We introduce the {\em robust} FRL-EH framework by formulating a novel global objective function. This function is designed to learn a global policy that achieves optimal worst-case performance across a predefined {\em covering set} of environments. This set formally encompasses the $K$ local environments, along with practically plausible perturbations of these environments.

    \item We propose FedRQ,  a tabular algorithm that learns a global policy within the robust FRL-EH framework. In contrast to QAvg, FedRQ enhances the local Q-learning update by incorporating a regularization term that reflects discrepancies across environments in the covering set. This design facilitates the learning of robust local Q functions that can effectively adapt to environmental variations.

    \item We provide a theoretical analysis showing that the global policy learned by FedRQ  converges asymptotically to the optimal solution of the proposed objective function. Specifically, the learned policy exhibits stable performance across both the local environments and their plausible perturbations.

    \item We extend FedRQ to environments with continuous state space. This extension introduces a key challenge: minimizing a value function over a large or continuous domain using only sampled data. To address this, we adopt the use of {\em expectile loss}, which enables the integration of FedRQ principles into various deep RL algorithms. In particular, we develop FedRDQN and FedRDDPG by incorporating our approach into DQN~\cite{mnih2015human} for discrete action spaces and DDPG~\cite{lillicrap2015continuous} for continuous action spaces, respectively.

    \item Through comprehensive experiments, we demonstrate that our FRL algorithms consistently outperform the current state-of-the-art methods such as DQNAvg and DDPGAvg~\cite{jin2022federated} across diverse heterogeneous environments. For fair comparison, we specifically focus on FedRDQN and FedRDDPG. Due to its scalability, our  approach can be seamlessly combined with other model-free RL algorithms.
\end{itemize}

\subsection{Outline}

The remaining part of this paper is organized as follows. Section~2 formally defines the FRL-EH framework and describes the de facto algorithm, QAvg. In Section~3, we propose FedRQ, a novel tabular algorithm designed to enhance robustness to environment heterogeneity within the FRL-EH framework. We also provide a theoretical proof of its asymptotic optimality. Section~4 extends FedRQ by incorporating its core ideas into widely-recognized RL algorithms, such as DQN and DDPG, resulting in FedRDQN and FedRDDPG for continuous state spaces. In Section 5, we evaluate the effectiveness of our proposed algorithms through experiments on FRL applications under heterogeneous environments. Finally, Section 6 concludes the paper.

\section{Preliminaries}\label{sec:Preliminaries}

We provide a formal definition of the Federated Reinforcement Learning (FRL), with a particular emphasis on scenarios characterized by heterogeneity in local environments. Building upon this, we introduce the de facto algorithm, known as QAvg. To enhance clarity in notation, we define $[N]:=\{1,2,...,N\}$ for any positive integer $N\geq 1$ throughout the paper.

\subsection{FRL with Environment Heterogeneity (FRL-EH)}

We begin by defining a Markov Decision Process (MDP), which serves as the mathematical foundation for Reinforcement Learning (RL) algorithms. An MDP is formally defined by a tuple $\Mc=\langle \Sc,\Ac, P, r, \gamma \rangle$, where $\Sc$ is the state space, $\Ac$ is the action space, $P:\mathcal{S}\times\mathcal{A}\rightarrow \Delta\mathcal{S}$ is the transition probability function, $r:\Sc \times \Ac \rightarrow [0, 1]$ is the reward function, and $\gamma \in (0,1)$ is the discount factor. Here, $\Delta \Sc$ denotes the probability simplex over the finite set $\Sc$ and $P(s' \mid s, a)$ represents the probability of transitioning to state $s'$ when action $a$ is taken at state $s$. At each time step $t$, an agent observes the current state $s_t \in \Sc$, selects an action $a_t \in \Ac$ according to its policy $\pi(\cdot \mid s_t)$, and receives the reward $r(s_t,a_t)$. Subsequently, the environment transitions to the next state $s_{t+1}$ according to the state transition probability distribution $P(\cdot \mid s_t, a_t)$.

We proceed to describe FRL, which comprises multiple environments.  Let $K$ denote the number of environments, each associated with a dedicated agent that interacts exclusively with its respective local environment. Specifically, for each $k \in [K]$, agent $k$ independently interacts with environment $k$, modeled as a MDP denoted by $\mathcal{M}_k$. For simplicity, we will refer to the environment $k$ as $\mathcal{M}_k$ throughout the paper. In the context of FRL, the primary objective is to collaboratively learn a global policy $\pi$ by leveraging the collective experience of all agents without requiring the sharing of raw data. It is typically assumed that all local environments are statistically identical~\cite{nadiger2019federated, han2019federated, wang2020federated}, implying that $\mathcal{M}_k$ is identical for all $k \in [K]$. However, this assumption is infrequently upheld in practice due to the inherent randomness and heterogeneity present across different environments.

To address this limitation, FRL with Environment Heterogeneity (FRL-EH) has been proposed \cite{jin2022federated}, which takes into account the discrepancies among local environments. In FRL-EH, it is assumed that environments may differ from one another, with environment heterogeneity primarily characterized by variations in state transition dynamics,  while sharing the state space $\mathcal{S}$, action space $\mathcal{A}$, and reward function $r$. Specifically, each environment $k$ is characterized by its respective MDP, defined as follows:
\begin{equation}\label{eq:Mk}
    \mathcal{M}_k = \langle \Sc, \Ac, P_k, r, \gamma \rangle,
\end{equation}
where $P_k: \mathcal{S} \times \mathcal{A} \times \mathcal{S} \rightarrow [0, 1]$ represents the unique state transition probability for environment $k \in [K]$.  Compared to conventional FRL, this approach provides a clearer understanding of how agents can operate effectively in diverse and dynamic environments by explicitly accounting for discrepancies across their local environments. This, in turn, can enhance the overall performance of the learned RL agents in practical scenarios.
Given a local environment $k \in [K]$, defined by $\Mc_k$, the {\em local} objective function for agent $k$ is defined as:
\begin{align}
    \Lc(\pi|P_k) &:= \mathbb{E} \Bigg[ 
        \sum_{t=1}^{\infty} \gamma^t r(s_t, a_t) \Big| \nonumber \\
        & s_0 \sim d_0, \ a_t \sim \pi(\cdot | s_t), \       s_{t+1} \sim P_k(\cdot | s_t, a_t) 
    \Bigg],
\end{align} where $d_0$ denotes a common initial distribution across the $K$ local environments.

\subsection{The De Facto Algorithm: QAvg}

In accordance with the standard framework of Federated Learning (FL), a global objective function for FRL-EH was introduced in \cite{jin2022federated}. This function is defined as the {\em average} of the $K$ local objective functions: 
\begin{equation}
    J_{\rm FRL}(\pi) = \frac{1}{K}\sum_{k=1}^{K}\Lc(\pi|P_k).
\end{equation} To optimize aforementioned global objective function within the FRL-EH framework, QAvg was proposed in \cite{jin2022federated} by integrating the Q-learning algorithm into FL. This method involves local updates performed by each agent, along with a global update executed by a central server. Specifically, each agent conducts a series of local updates based on the standard Q-learning algorithm. After completing a predefined number of local updates, the agents transmit their updated Q functions (commonly referred to as local updates) to the central server. The server then performs a global update by averaging the received Q functions in a manner analogous to the standard approach used in FL. The averaged Q function (commonly referred to as a global update) is subsequently broadcast back to the $K$ agents. The local and global update procedures in QAvg are outlined as follows:

\vspace{0.1cm}
\noindent{\bf Local update:} At each time step $t$, each agent $k \in [K]$ performs a local update according to the principle of Q-learning \cite{watkins1992q}, which is defined as follows:
\begin{align}\label{eq:QAvgLocal}
    & Q^k_{t+1}  (s, a)  \leftarrow  (1 - \lambda_t)Q^k_t(s, a) \nonumber \\ 
    & \quad + \lambda_t \left[ r(s, a) + \gamma\sum_{s'}P_k(s'\mid s, a) \max_{a'\in\mathcal{A}} Q^k_t(s', a') \right],
\end{align} for all $(s,a) \in \mathcal{S} \times \mathcal{A}$, where $Q^k_t$ denotes the action-value function (a.k.a. Q function) of agent $k$, $(s, a, s') \in \mathcal{S} \times \mathcal{A} \times \mathcal{S}$ represent the current state, action, and next state, respectively. The parameter $\lambda_t$ denotes the learning rate.

\vspace{0.1cm}
\noindent{\bf Global update:} Let $\mathcal{I}_E := \{nE \mid  n \in \mathbb{N} \}$, where $E$ represents the period of the global update (or the number of local updates per each global update). At each time step $t \in \mathcal{I}_E$, the central server updates a global Q function, denoted as $\bar{Q}_t$, by averaging the aggregated local Q functions.  Specifically, the global update of QAvg is defined as follows:
\begin{align}\label{eq:QAvgGlobal}
    \bar{Q}_{t+1}(s,a) \leftarrow \frac{1}{K} \sum_{k=1}^{K} Q_{t+1}^{k}(s,a),\;\;\; \forall (s,a) \in \Sc \times \Ac.
\end{align} 
After global aggregation, the central server broadcasts the updated global function $\bar{Q}_{t+1}$ to all $K$ agents. Consequently, the local Q function of agent $k \in [K]$ is updated as follows:
\begin{equation}\label{eq:QAvgBroadcast}
    {Q}^k_{t+1}(s, a) \leftarrow \bar{Q}_{t+1}(s, a),\;\; \forall (s,a)\in \Sc\times\Ac.
\end{equation}

In this process, only the Q functions are exchanged between the agents and the central server, while the individual experience data of each agent remains confidential. This design aligns with the fundamental principles of FL, ensuring that data privacy is preserved throughout the learning procedure. 

By incorporating function approximation methods such as Deep Q-Network (DQN)~\cite{mnih2013playing} and Deep Deterministic Policy Gradient (DDPG)~\cite{fujimoto2018addressing}, QAvg has been extended to {\bf DQNAvg} for discrete action space environments and {\bf DDPGAvg} for continuous action space environments. This extension facilitates he management of large-scale and complex environments. These methods will serve as benchmark algorithms in our experimental evaluations.

\begin{remark} (Convergence of QAvg) In  \cite{jin2022federated}, it was theoretically proved that QAvg asymptotically converges to a suboptimal solution of $J_{\rm FRL}(\pi)$. More precisely, this corresponds to the optimal solution of the following objective function, which serves as the lower bound of $J_{\rm FRL}(\pi)$: 
\begin{equation}
    J_{\rm Lower}(\pi)=\Lc\left(\pi\Big|\bar{P}=\frac{1}{K}\sum_{k=1}^{K}P_k\right).\label{eq:QAvgGlobalObjective}
\end{equation} The policy derived from QAvg can asymptotically converge to an optimal policy of a virtual MDP, defined as $\langle \Sc, \Ac, \bar{P}, r, \gamma \rangle$. It can be reasonably expected that QAvg will exhibit favorable performance when $P_k$'s are concentrated around the average transition dynamic $\bar{P}$. However, as the heterogneity among the local environments increases, QAvg may struggle to yield satisfactory performance.
\end{remark}

\section{Main Idea}

To more accurately reflect practical scenarios, we propose a {\em robust} FRL-EH framework, which incorporates a novel global objective function aimed at optimizing a global policy that ensures stable performance across all $K$ environments, as well as in potentially perturbed environments. As a result, this framework enables our learned policy to operate robustly in diverse and heterogeneous environments, extending beyond the local environments.

To this end, we introduce a {\em covering set} $\Pc$ designed to capture the heterogeneity across environments. Based on this set, we formally define the proposed global objective function. Furthermore, we develop a tabular algorithm named FedRQ, which demonstrates that the policy learned through FedRQ can converge to the optimal solution of our global objective function as the round of federation (i.e., global updates) increases.

\subsection{Robust FRL-EH Framework}

To delineate the proposed robust FRL-EH framework, we first introduce a covering set $\Pc$ that satisfies the condition:
\begin{equation}\label{eq:cond}
    \{P_k: k \in [K]\} \subseteq \Pc,
\end{equation} where $P_k$, defined in \eqref{eq:Mk}, denotes the transition probability distribution for environment $k \in [K]$. This set accounts for potential variations in transition dynamics across environments. Optimizing a policy over this covering set can yield a global policy that ensures consistent performance across the $K$ local environments as well as in potentially perturbed environments, provided that $\Pc$ is appropriately designed.
Accordingly, given a predefined covering set $\Pc$, our proposed global objective function is defined as follows:
\begin{equation}\label{eq:FRLEHObjective1}
   J^{\rm FRL}(\pi) := \inf_{P \in \mathcal{P}} \; \Lc(\pi|P).
\end{equation} Rather than focusing on average performance over the $K$ local environments,  this function emphasizes the worst-case performance across all plausible environments, thereby ensuring the {\em robustness} of the learned agent. Consequently, our robust FRL-EH framework is particularly well-suited for safety-critical RL applications,  such as autonomous driving and robotics scenarios \cite{gu2024review}.

It is important to note that if a covering set $\mathcal{P}$ is poorly constructed---whether too narrow or excessively broad---the resulting policy may fail to generalize effectively across diverse environments. While selecting the minimal set $\mathcal{P} = \{ P_k: k \in [K]\}$ guarantees stable performance across the $K$ local environments, it may struggle with perturbed environments that are plausible in practice. Conversely, if $\mathcal{P}$ is defined as the universal set encompassing all possible state transition probability distributions, it becomes infeasible to identify an optimal policy that satisfies the global objective function due to the excessively large and uninformative nature of the covering set. Therefore, the meticulous design of the covering set is crucial to the efficacy of the robust FRL-EH framework.

From now on, we present an efficient and practical covering set. To achieve this, we define the average of the state transition probabilities across all $K$ environments as follows:
\begin{equation}\label{eq:AverageStateTransitionProbability}
    \bar{P}(\cdot|s,a):=\frac{1}{K}\sum_{k=1}^{K}P_k(\cdot|s,a),\; \forall (s,a)\in \Sc\times \Ac.
\end{equation} For each state $s \in \Sc$, we define its neighboring set as:
\begin{equation} \label{eq:NeighboringSet}
\mathcal{N}^s := \left\{ s^\prime \in \mathcal{S}\; \middle\vert \; \sum_{a \in \mathcal{A}}\bar{P}(s'|s,a) \ne 0 \right\},
\end{equation} which comprises the possible next states from the current state $s$. Leveraging this, we define the set of all feasible conditional distributions at state $s$ as follows: 
\begin{equation}
\mathcal{Q}^s = \left\{\qv \in \Delta_{|\mathcal{S}|} \mid \qv({s^{\prime}}) = 0, \; \forall s' \in \mathcal{S} -  \mathcal{N}^s \right\} \subseteq \Delta_{|\mathcal{S}|}.
\end{equation}  This set constraints the possible state transition dynamics within a localized neighborhood, thereby preventing transitions to unrealistic states and ensuring that only realistic transition dynamics are included in the covering set. Building upon this set, we define the set of perturbed probability distributions centered the average distribution $\bar{P}(\cdot|s,a)$ with a robustness (or covering) level $\omega \in [0,1)$: 
\begin{equation}
    \Pc^{(s,a)}_{\omega} = \left\{(1-\omega)\bar{P}(\cdot|s,a)+\omega \qv \mid  \qv \in \mathcal{Q}^s\right\}.\label{eq:subset}
\end{equation} Utilizing the Kronecker product of $\{\Pc^{(s,a)}: (s,a)\in\Sc\times \Ac\}$, our proposed covering set is defined as:
\begin{equation}\label{eq:AdjacentUncertaintySet}
\mathcal{P}_{\omega} := \bigotimes_{(s,a)\in\mathcal{S}\times \mathcal{A}}\mathcal{P}^{(s,a)}_{\omega}.
\end{equation} 

We will show that by appropriately selecting the robustness level $\omega$, our covering set fulfills the condition in \eqref{eq:cond}, which implies that 
\begin{equation}
    P_{k}(\cdot|s,a)\in \Pc_{\omega}^{(s,a)}, \quad \forall (s,a)\in\Sc\times \Ac.
\end{equation}
To achieve this, the covering level $\omega$ must meet the following requirement:
\begin{equation}
    \bar{P}(\cdot|s,a) + \frac{1}{\omega}\left(P_k(\cdot|s,a) - \bar{P}(\cdot|s,a)\right) \in \Delta_{|\Sc|}, \forall k \in [K],
\end{equation} which can be equivalently expressed as: 
\begin{equation}
    \omega+\frac{P_k(s'|s,a)-\bar{P}(s'|s,a)}{\bar{P}(s'|s,a)}\geq 0,
\end{equation}
for all $s' \in \Sc$ and $k \in [K]$. By leveraging this, we can identify that the condition $ P_k(\cdot|s,a) \in \Pc_{\omega}^{(s,a)}, \forall k \in [K]$ holds, given that $\omega$ is selected to satisfy:
\begin{equation}\label{eq:omega}
    \omega \geq \max_{(s,a)\in\Sc\times\Ac}\kappa(s,a).
\end{equation} Here, $\kappa(s,a)$ represents the heterogeneity (or concentration) of the $K$ local environments. Therefore, we can ensure that 
\begin{equation}
    P_k \in \Pc_{\omega}, \forall k \in [K],
\end{equation} provided that $\omega$ is chosen such that
\begin{equation}\label{eq:omega}
    \omega \geq \max_{(s,a)\in\Sc\times\Ac}\kappa(s,a).
\end{equation}
It follows that a larger robustness level $\omega$ must be selected as the heterogeneity of the local environment increases.  In the case of homogeneous environments, it is natural to select $\omega=0$.

Hereafter, our global objective function, as presented in \eqref{eq:FRLEHObjective1}, is assumed to incorporate the proposed covering set $\Pc_{\omega}$ as defined in \eqref{eq:AdjacentUncertaintySet}. Under this global objective function, the state value function, denoted as $V_{R}^{\pi}$ and the action value function (or Q function), denoted as $Q_{R}^{\pi}$ for a policy $\pi$ are defined as:
\begin{align}
    V_{R}^{\pi}(s) &:= \inf_{P\in \Pc_{\omega}}\EE_{\pi, P}\left[\sum_{t=0}^{\infty} \gamma^t r(s_t,a_t)\middle\vert s_0=s \right]\label{eq:V}\\
    Q_{R}^{\pi}(s,a)&:=\inf_{P\in \Pc_{\omega}}\EE_{\pi, P}\left[\sum_{t=0}^{\infty} \gamma^t r(s_t,a_t)\middle\vert s_0=s, a_0=a \right].\label{eq:Q}
\end{align}  The optimal policy $\pi_{R}^{\star}$, which represents the optimal solution of our global objective function in \eqref{eq:FRLEHObjective1}, is derived as:
\begin{equation}
    \pi_{R}^{\star}=\argmax_{\pi} V_{R}^{\pi}(s),\;\; \forall s \in \Sc. 
\end{equation} Furthermore, the optimal value functions are defined as:
\begin{align}
    V_{R}^{\star}:= V_{R}^{\pi_{R}^{\star}}\; \mbox{and }\; Q_{R}^{\star}:= Q_{R}^{\pi_{R}^{\star}}.\label{eq:Qoptimal}
\end{align}

\subsection{The Proposed FedRQ}

Our primary objective is to derive a global Q function, denoted as $\bar{Q}_{t}$, that converges to the optimal Q function, $Q_{R}^{\star}$, as the global round $t$ increases. This convergence is essential for ensuring that the policy closely approximates optimal decision-making within the proposed robust framework for FRL-EH. To achieve this, we present a novel FRL-EH algorithm, named FedRQ. While the standard QAvg attempts to address the heterogeneity solely though averaging during the global update, our proposed method further incorporates a regularization term during local updates. Specifically, FedRQ utilizes a {\em robust} local update mechanism, which is articulated as follows:

\vspace{0.1cm}
\noindent{\bf Robust local update:} At each time step $t$, each agent $k \in [K]$ performs the local update following the {\em robust} Q-learning principle \cite{hwang2024practical}, which is defined as:
\begin{align}\label{eq:RedRQLocal}
       & Q^k_{t+1}  (s, a)  \leftarrow  (1 - \lambda_t)Q^k_t(s, a)   \nonumber \\ 
    &  + \lambda_t \Bigg[ r(s, a) + \gamma (1-\omega) \sum_{s' \in \mathcal{S}}P_k(s'\mid s, a)\max_{a'\in\mathcal{A}}Q^k_t(s', a')  \nonumber \\
    &  + \gamma \omega \min_{s' \in \mathcal{N}^s_k}\max _{a' \in \mathcal{A}}Q^k_t(s', a') \Bigg],\; \forall (s,a)\in\Sc\times \Ac,
\end{align} where $Q^k_t$ denotes the Q function of agent $k$ at time step $t$, $(s, a, s') \in \mathcal{S} \times \mathcal{A} \times \mathcal{S}$ represent the current state, action, and next state, respectively. Additionally, $\lambda_t$ is the learning rate, $\omega$ indicates the robustness level, and $\mathcal{N}^s_k$ denotes the neighboring set of state $s \in \mathcal{S}$ for the local environment $k$, defined as:
\begin{equation}\label{eq:NeighboringSetK}
    \Nc^s_{k} := \left\{ s^\prime \in \mathcal{S}\; \middle\vert \; \sum_{a \in \mathcal{A}}P_k(s' \mid s, a) \ne 0 \right\}.
\end{equation}

\vspace{0.1cm}
\noindent{\bf Global update:} Let $\mathcal{I}_E := \{nE \mid  n \in \mathbb{N} \}$, where $E$ represents the period of the global update (or the number of local updates per each global update). At each time step $t \in \mathcal{I}_E$, the central server updates a global Q function, denoted as $\bar{Q}_t$, by averaging the aggregated local Q functions. Specifically, the global update of FedRQ is defined as:
\begin{align}\label{eq:FedRQGlobal}
    \bar{Q}_{t+1}(s,a) \leftarrow \frac{1}{K} \sum_{k=1}^{K} Q_{t+1}^{k}(s,a),\quad \forall (s,a)\in \Sc\times \Ac.
\end{align} 
After global aggregation, the central server disseminates the updated global model $\bar{Q}_{t+1}$ to the $K$ agents. Consequently, the local Q-function of agent $k \in [K]$ is updated as follows:
\begin{equation}\label{eq:QAvgBroadcast}
    {Q}^k_{t+1}(s, a) \leftarrow \bar{Q}_{t+1}(s, a),\quad \forall (s,a) \in \Sc\times\Ac.
\end{equation} Similar to QAvg, only the Q-functions are exchanged among agents during the execution of FedRQ, while each agent's individual experiences remain private.

Notably, our proposed local update incorporates the additional term:
\begin{equation}\label{eq:PerturbationTerm}
    \min_{s' \in \Nc_{P_k}^{s}} \max_{a' \in \Ac} Q_{t}^k(s',a').
\end{equation} 
This term plays a crucial role in mitigating potential risks arising from environmental heterogeneity. Each local agent trains its local policy under the assumption that variations may exist in the environments of other agents, as modeled in \eqref{eq:PerturbationTerm}. By incorporating this consideration, our local update mechanism produces a Q function that exhibits improved robustness against such heterogeneity.
Furthermore, Theorem~\ref{theorem:theorem1} in Section~\ref{subsec:TA} substantiates the optimality of our proposed FedRQ, confirming that $\bar{Q}_{t+1}$, as defined in \eqref{eq:FedRQGlobal}, converges to the optimal Q-function $Q_{R}^{\star}$ as the global round $t$ increases.

Our proposed FedRQ significantly enhances the learning process within the FRL-EH framework by incorporating a regularization term in the local update, which effectively addresses environmental heterogeneity alongside global (or federated) averaging. In contrast, QAvg merely takes the average during the global update to tackle this challenge. This key distinction enables FedRQ to improve the robustness of the learned global Q function by comprehensively accounting for variations in local environments, while QAvg relies on a more straightforward averaging approach. Notably, QAvg can be viewed as a special case of FedRQ with $\omega = 0$.

\subsection{Theoretical Analysis}\label{subsec:TA}

We theoretically prove that the global Q function, derived from our proposed FedRQ, converges to the optimal Q function $Q_{R}^{\star}$, which is associated with our global objective function, as defined in \eqref{eq:Qoptimal}. To ensure the comprehensiveness of our analysis, we present the following assumption:

\vspace{0.1cm}
\begin{assumption}\label{assumption:assumption1} (Consistency of non-zero transitions) For every $s \in \mathcal{S}$, it holds that
 \begin{equation*}
        \mathcal{N}^s_{1} = \mathcal{N}^s_{2}=\cdots = \mathcal{N}^s_K.
\end{equation*}
\end{assumption}
This assumption is valid when the set of states with non-zero transition probabilities from state $s$ remains consistent across all local environments, despite variations in the corresponding transition probabilities. In many real-world applications, it is reasonable to expect identical prohibitive transitions, even among local environments with heterogeneity.

The main result of this section concerning the convergence analysis of our FedRQ is presented below:

\begin{theorem}(Convergence of FedRQ)\label{theorem:theorem1}
Let $\lambda_t = \frac{2}{(1 - \gamma)(t + E)}$, where $\gamma\in [0.2, 1)$ and $E>1$ denotes the period of the global update.  Under Assumption 1, the global Q function, defined as $\bar{Q}_t = \frac{1}{K}\sum_{k=1}^{K} Q_t^k$, converges to the optimal Q function $Q_R^\star$ as defined in~\eqref{eq:Qoptimal}, with a vanishing gap:
\begin{equation}
    \left\Vert \bar{Q}_t - Q_R^\star \right\Vert_\infty \le \frac{16\gamma (E-1) }{(1 - \gamma)^3 (t + E)}.
\end{equation}
\end{theorem}
\begin{IEEEproof}
We first define the robust Bellman operator for the MDP $\Mc_{k}$, which corresponds to the environment $k$, as follows:
\begin{align}
    \mathcal{T}_k Q(s, a) & := r(s, a) \nonumber \\
    & \quad + \gamma \Bigg[(1 - \omega) \sum_{s' \in \mathcal{S}} P_k(s' \mid s, a) \max_{a' \in \mathcal{A}} Q(s', a') \nonumber \\
    & \quad\quad + \omega \min_{s' \in \mathcal{N}^s_{P_k}} \max_{a' \in \mathcal{A}} Q(s', a') \Bigg].
\end{align}
Similarly, the robust Bellman operator corresponding to the global objective function defined in~\eqref{eq:FRLEHObjective1} is given as follows:
\begin{align}
    \bar{\mathcal{T}} Q(s, a) & := r(s, a) \nonumber \\
    & \quad + \gamma \Bigg[(1 - \omega) \sum_{s' \in \mathcal{S}} \bar{P}(s' \mid s, a) \max_{a' \in \mathcal{A}} Q(s', a') \nonumber \\
    & \quad\quad + \omega \min_{s' \in \mathcal{N}^s_{\bar{P}}} \max_{a'\in\mathcal{A}} Q(s', a') \Bigg],
\end{align} where $\bar{P}$ denotes the average state transition probability defined in \eqref{eq:AverageStateTransitionProbability}. Under Assumption 1, the following relationship holds:
\begin{align}
    \bar{\mathcal{T}} Q = \frac{1}{K} \sum_{k=1}^K \mathcal{T}_k Q.\label{eq:relationship}
\end{align} 
As proven in~\cite{hwang2024practical}, since $\bar{\mathcal{T}}$ is a $\gamma$-contraction mapping, there exists a unique fixed point $Q_{R}^{\star}$ such that $\bar{\mathcal{T}} Q_{R}^{\star} = Q_{R}^{\star}$.

In the FedRQ, the update rule can be expressed as follows. Each agent $k$ performs the local update:
\begin{align}
    Q^k_{t+1} & \leftarrow (1 - \lambda_t) Q^k_t + \lambda_t \mathcal{T}_k Q^k_t.\label{eq:LU}
\end{align}
At synchronization step $t \in \mathcal{I}_E$,  all local Q functions are globally updated via averaging:
\begin{align}
    Q^k_{t+1} & = 
    \begin{cases} 
        \frac{1}{K} \sum_{k=1}^K Q^k_{t+1}, & \text{if } t \in \mathcal{I}_E, \\
        Q^k_{t+1}, & \text{otherwise}.
    \end{cases} \label{eq:intermittent}
\end{align}

We are now ready to prove Theorem 1. From the supporting lemmas (Lemma 1 and Lemma 2 in the last of this proof), we derive the following recursive inequality:
\begin{align}
    \left\| \bar{Q}_{t+1} - Q_{R}^{\star} \right\|_\infty 
    \le & \left(1 - (1 - \gamma)\lambda_t \right) \left\| \bar{Q}_t - Q_{R}^{\star} \right\|_\infty \nonumber \\
    & + \frac{4\gamma \lambda_t^2(E-1) }{1 - \gamma}.
\end{align} Letting $\Delta_t := \left\| \bar{Q}_t - Q_{R}^{\star} \right\|_\infty$, the above inequality can be rewritten as:
\begin{align}
    \Delta_{t+1} &\le \left(1 - (1 - \gamma)\lambda_t \right) \Delta_t + \frac{4\gamma \lambda_t^2 (E-1) }{1 - \gamma}\\
    &\stackrel{(a)}{=}\left(\frac{t+E-2}{t+E}\right)\Delta_{t} + \frac{\zeta}{(t+E)^2}, \label{eq:E1}
\end{align} where $\zeta := \frac{16\gamma (E-1) }{(1 - \gamma)^3}$ and (a) is due to the choice of $\lambda_t = \frac{2}{(1-\gamma)(t+E)}$. Suppose that the following upper bound holds: 
\begin{equation}
    \Delta_t \le \frac{\zeta}{t + E}=\frac{16\gamma (E-1)}{(1-\gamma)^3(t+E)}. \label{eq:E2}
\end{equation} Then, we can show that 
\begin{align}
    \Delta_{t+1} 
    &\stackrel{(a)}{\leq} \left(\frac{t+E-2}{t+E}\right)\left(\frac{\zeta}{t + E}\right) + \frac{\zeta}{(t+E)^2}\\
    &\stackrel{(b)}{\leq} \frac{\zeta}{t + E + 1}.\label{eq:E3}
\end{align} where (a) follows from \eqref{eq:E1} and \eqref{eq:E2}, and (b) is due to the following fact:
\begin{equation*}
    \frac{1}{t+E} - \frac{1}{t+E+1} \leq \frac{1}{(t+E)^2}.
\end{equation*} Finally, in the case of $t=1$, the inequality in~\eqref{eq:E2} holds as:
\begin{align}
    \Delta_1 \le \frac{1}{1 - \gamma} \stackrel{(a)}{\le} \frac{\zeta}{1 + E} 
    = \frac{16\gamma(E - 1)}{(1 - \gamma)^3(1 + E)},
\end{align} for $E>1$. Since $\frac{E-1}{1+E}\geq \frac{1}{3}$, the above inequality is satisfied as long as $\gamma \geq 0.15$ (i.e., $\gamma \in [0.2, 1)$). Thus, by combining \eqref{eq:E2} and \eqref{eq:E3}, we complete the proof of Theorem 1.
\end{IEEEproof}

\begin{lemma}\label{lemma:lemma1}
For any round $t\geq 1$, the following upper bound holds:
\begin{align}
    \left\| \bar{Q}_{t+1} - Q_{R}^{\star} \right\|_\infty \le & \left(1 - (1 - \gamma)\lambda_t \right) \left\| \bar{Q}_t - Q_{R}^{\star} \right\|_\infty \nonumber \\
    & + \frac{\gamma \lambda_t}{K} \sum_{k=1}^K \left\| Q^k_t - \bar{Q}_t \right\|_\infty.
\end{align}
\begin{IEEEproof}
    Leveraging the relationship in \eqref{eq:relationship} and the local update in \eqref{eq:LU}, we can get:
    \begin{align*}
    & \left\| \bar{Q}_{t+1} - Q_{R}^{\star} \right\|_\infty \\
    &= \left\| (1 - \lambda_t)\bar{Q}_t + \frac{\lambda_t}{K}\sum_{k=1}^K \mathcal{T}_k Q^k_t - Q_{R}^{\star} \right\|_\infty \\
    &\stackrel{(a)}{=} \left\| (1 - \lambda_t)(\bar{Q}_t - Q_{R}^{\star}) + \frac{\lambda_t}{K}\sum_{k=1}^K (\mathcal{T}_k Q^k_t - \mathcal{T}_k Q_{R}^{\star}) \right\|_\infty \\
    &\leq (1 - \lambda_t)\left\| \bar{Q}_t - Q_{R}^{\star} \right\|_\infty + \frac{\lambda_t}{K} \sum_{k=1}^K \left\| \mathcal{T}_k Q^k_t - \mathcal{T}_k Q_{R}^{\star} \right\|_\infty \\
    &\stackrel{(b)}{\leq} (1 - \lambda_t)\left\| \bar{Q}_t - Q_{R}^{\star} \right\|_\infty + \frac{\gamma \lambda_t}{K} \sum_{k=1}^K \left\| Q^k_t - Q_{R}^{\star} \right\|_\infty \\
    &\leq (1 - \lambda_t)\left\| \bar{Q}_t - Q_{R}^{\star} \right\|_\infty + \frac{\gamma \lambda_t}{K} \sum_{k=1}^K \left\| Q^k_t - \bar{Q}_t \right\| \nonumber\\
    &\quad\quad\quad + \gamma\lambda_t\left\| \bar{Q}_t - Q_{R}^{\star} \right\|_\infty \\
    & = \left(1 - (1 - \gamma)\lambda_t \right) \left\| \bar{Q}_t - Q_{R}^{\star} \right\|_\infty + \frac{\gamma \lambda_t}{K} \sum_{k=1}^K \left\| Q^k_t - \bar{Q}_t \right\|_\infty,
\end{align*} where (a) is due to the fact that $\Tc_{k} Q_{R}^{\star} = Q_{R}^{\star}$ and (b) is because $\Tc_k$ is $\gamma$-contractor~\cite{hwang2024practical}. This completes the proof of Lemma 1.
\end{IEEEproof}
\end{lemma}

\begin{lemma} Let $\lambda_t = \frac{2}{(1-\gamma)(t+E)}$. The following upper bound holds:
\begin{align}
    \frac{1}{K} \sum_{k=1}^K \left\| Q^k_t - \bar{Q}_t \right\|_\infty \le \frac{4 \lambda_t (E - 1)}{1 - \gamma},
\end{align} where $\lambda_t = \frac{2}{(1-\gamma)(t+E)}$.
\begin{IEEEproof}
For every time step $t$, there exists some $t_0 \in [t - E + 1, t]$ such that $Q^k_{t_0} = \bar{Q}_{t_0}$ due to the intermittent global update, as outlined in \eqref{eq:intermittent}. Namely, at time step $t_0$, the synchronization step occurs.
Accordingly, the following inequality holds:
\begin{align*}
    & \frac{1}{K} \sum_{k=1}^K \left\| Q^k_t - \bar{Q}_t \right\|_\infty \\
    &= \frac{1}{K} \sum_{k=1}^K \left\| Q^k_t - \bar{Q}_{t_0} + \bar{Q}_{t_0} - \bar{Q}_t \right\|_\infty \\
    &{\le} \frac{2}{K} \sum_{k=1}^K \left\| Q^k_t - \bar{Q}_{t_0} \right\|_\infty \\
    &\stackrel{(a)}{=} \frac{2}{K} \sum_{k=1}^K \left\| \sum_{t' = t_0}^{t - 1} \lambda_{t'} \left( \mathcal{T}_k Q^k_{t'} - Q^k_{t'} \right) \right\|_\infty \\
    &\le \frac{2}{K} \sum_{k=1}^K \sum_{t' = t_0}^{t - 1} \lambda_{t'} \left\| \mathcal{T}_k Q^k_{t'} - Q^k_{t'} \right\|_\infty \\
    & \stackrel{(b)}{\le} 2 \sum_{t' = t_0}^{t - 1} \lambda_{t'} \frac{1}{1 - \gamma} \\
    &\stackrel{(c)}{\leq} \frac{4 \lambda_t (E-1)}{1 - \gamma},
\end{align*}where (a) follows from the local update rule in \eqref{eq:LU}, (b) is due to the fact that $Q^{k}_{t}(s,a)\leq\frac{1}{1-r}$, and  (c) is obtained from $\lambda_{t'} \leq \lambda_{t}$ for  $t' \in [t_0 , t-1]$ with the choice of $\lambda_t = \frac{2}{(1-\gamma)(t+E)}$. This completes the proof of Lemma 2.   
\end{IEEEproof}
\end{lemma}

\section{Practical FRL Algorithms with Function Approximations}

We extend the proposed FedRQ algorithm to the function approximation setting in order to effectively address practical scenarios involving environments characterized by large or continuous state and action spaces. This extension is essential, as real-world applications often operate within high-dimensional environments that require advanced methods for efficient learning and decision-making. In contrast to the QAvg algorithm~\cite{jin2022federated}, a key challenge in this extension lies in the minimization over neighboring states, as described in~\eqref{eq:RedRQLocal}. The exhaustive search over this set becomes impractical in real-world scenarios involving large or continuous state spaces. To address this challenge, we introduce a fundamental concept based on expectile loss in Section~\ref{subsec:DegreeFunction}. Furthermore, in Sections~\ref{subsec:FedRDQN} and~\ref{subsec:FedRDDPG}, we develop various FRL algorithms with function approximation as extensions of FedRQ.

Throughout the paper, we denote the replay buffer for agent $k$ as
$\Dc_k=\{(s=s_t,a=a_t,r=r(s_t,a_t),s'=s_{t+1})\}$. This buffer comprises a collection of samples obtained through the interactions of agent $k \in [K]$ with its respective local environment.

\subsection{Expectile Network}\label{subsec:DegreeFunction}

We aim to efficiently approximate the minimization over the neighboring set in~\eqref{eq:RedRQLocal} using samples drawn from the replay buffer. This approach is particularly crucial in environments with large or continuous state spaces, where exhaustively searching the neighboring set is computationally impractical. To address this challenge, given a Q function $Q(\cdot,\cdot)$, we first introduce the {\em degree function} for environment $k$, defined as follows:
\begin{equation}\label{eq:FRLDetector}
    D_k(s) := \min_{s' \in \mathcal{N}^s_{k}} \max_{a' \in \mathcal{A}}\;  Q(s', a').
\end{equation} This degree function is approximated using a deep neural network (DNN) parameterized by $\psi_k$ (referred to as the {\em expectile network}), and is trained using the following expectile regression-based loss:
\begin{equation}\label{eq:FRLDetectorLoss}
    \Lc_{D}(\psi_k) = \mathbb{E}_{(s, s') \sim \mathcal{D}_k} \left[ \ell_\tau\left(\max_{a'\in\mathcal{A}}\; Q(s', a'), D(s ; \psi_k)\right) \right],
\end{equation}
where $\ell_\tau$ denotes the expectile loss function \cite{newey1987asymmetric}, defined as:
\begin{equation} \label{eq:ExpectileRegressionLoss}
    \ell_\tau(y, x) = 
    \begin{cases}
        \tau(y - x)^2          & \text{if} \quad y \ge x    \\
        (1-\tau)(y - x)^2      & \text{if} \quad y < x   
    \end{cases},
\end{equation}
and $\tau \in (0, 0.5)$ is the expectile level that emphasizes the lower tail of the distribution. 
In this paper, we set $\tau = 0.01$, which allows the model to place greater emphasis on the minimum among the sampled values of $\max_{a' \in \mathcal{A}} Q(s', a')$. Moreover, the sample pairs $(s, s')$ from $\mathcal{D}_k$ are constructed such that $s' \in \mathcal{N}^s_{k}$. 
As a result, the loss function in~\eqref{eq:FRLDetectorLoss} enables the degree function to efficiently approximate the minimization over the neighboring set using standard gradient-based methods with samples drawn from the replay buffer.

\begin{algorithm}[h!]
    \caption{FedRDQN} \label{alg:FedRDQNAlgorithm}
    \begin{algorithmic}[1] 
        \STATE {\bf Input:} The discount factor $\gamma \in (0, 1)$, learning rates $\lambda_V > 0$, target update rate $\eta \in (0, 1)$, mini-batch size $B$, exploration rate $\epsilon \in [0, 1]$, global update interval $E$, total time steps $T$, update starting time step $T_u$, robustness level $\omega \in [0, 1]$, and expectile level $\tau = 0.01$.

        \STATE {\bf Initialization:} For all agents $k \in [K]$, initialize network parameters $\theta_k, \psi_k$, set target network parameters as $\theta'_k \leftarrow \theta_k$, initialize the replay buffer $\mathcal{D}_k = \emptyset$, and start with the initial state $s_0^k \in \mathcal{S}$.

        \FOR{$t=0$ to $T-1$}
            \FOR{each $k \in [K]$}
                \STATE With probability $\epsilon$, select a random action $a_t^k \in \mathcal{A}$
                \STATE Otherwise, $a_t^k = \argmax_{a \in \mathcal{A}} Q(s_t^k, a; \theta_k)$
                \STATE \quad\quad\quad\quad\quad $s^k_{t+1} \sim P_k(\cdot \mid s^k_t, a^k_t)$
                \STATE $\mathcal{D}_k \leftarrow \mathcal{D}_k \cup \{ (s^k_t ,a^k_t, r(s^k_t,a^k_t), s^k_{t+1} \}$
                
                \IF{$t \ge T_u$}
                    \STATE Sample $(s, a, r, s') \sim \mathcal{D}_k$
                    \STATE Update local parameters $\theta_k$ and $\psi_k$ via \eqref{eq:FedRDQNQLocal} and \eqref{eq:FedRDQNExpectileLocal}, respectively
                    \STATE Update target parameters $\theta'_k$ via $\eqref{eq:FedRDQNTargetLocal}$
                \ENDIF
                \IF{The environment is truncated or done}
                    \STATE Reset the environment
                \ENDIF
            \ENDFOR
            \IF{$t + 1 \in \mathcal{I}_E$}
                \STATE Update the global parameters $(\bar{\theta}, \bar{\theta}', \bar{\psi})$ via \eqref{eq:FedRDQNGlobal}
                \STATE Broadcast the updated global parameters $(\bar{\theta}, \bar{\theta}', \bar{\psi})$ to all $K$ agents
                \STATE Update the local parameters $(\theta_k, \theta'_k, \psi_k) \leftarrow (\bar{\theta}, \bar{\theta'}, \bar{\psi})$ for each agent $k \in [K]$
            \ENDIF
        \ENDFOR
    \end{algorithmic}
\end{algorithm}

\subsection{FedRDQN}\label{subsec:FedRDQN}

DQN is a widely adopted function approximation algorithm in RL, particularly for environments with discrete action spaces. It leverages a DNN to approximate the Q function, enabling a more expressive representation compared to the traditional tabular Q-learning. To enhance training stability, DQN employs a separate target network to compute the temporal-difference (TD) target. This target network is updated gradually to align with the main network, thereby stabilizing the learning process.

FedRDQN extends the proposed FedRQ algorithm by incorporating the DQN architecture. Specifically, it leverages DNNs to approximate the local Q functions and their corresponding target functions. Furthermore, FedRDQN incorporates the degree function defined in~\eqref{eq:FRLDetector} to effectively capture environmental discrepancies during the learning process. Based on this, the loss function for the Q function in environment $k$ is defined as:
\begin{equation}\label{eq:FedRDQNQLoss}
    \Lc_{Q}(\theta_k)  = \mathbb{E}_{(s, a, r, s') \sim \mathcal{D}_k}\Bigg[ \Big( Q(s, a; \theta_k) - y \Big)^2 \Bigg],
\end{equation} where
\begin{equation}
    y  = r + \gamma (1-\omega)  \max_{a' \in \mathcal{A}}\; Q  \left(s', a'; \theta'_k \right)   + \gamma\omega D(s ; \psi_k).
\end{equation} Herein, $\theta_k$ and $\theta'_k$ denote the parameters of the Q network and its corresponding target network, respectively. Additionally, $\psi_k$ represents the parameters of the expectile network and $\omega$ denotes a predefined robustness level. 

It is important to note that the loss function in~\eqref{eq:FedRDQNQLoss} can be computed solely using samples collected from the local environment $k$. This property is crucial for incorporating model-free (i.e., sample-based) RL algorithms into our proposed FedRQ framework. Now, we formally define the local and global update processes in FedRDQN.

\vspace{0.1cm}
\noindent\textbf{Local Update:} Each agent $k \in [K]$ updates its local parameters at each time step $t$. Specifically, the parameters of the Q network and the expectile network are updated with a learning rate $\lambda_V$ as:
\begin{align} 
    \theta_k &\leftarrow \theta_k - \lambda_{V} \nabla \mathcal{L}_Q(\theta_k)\label{eq:FedRDQNQLocal}\\
    \psi_k &\leftarrow \psi_k - \lambda_{V} \nabla \mathcal{L}_D(\psi_k), \label{eq:FedRDQNExpectileLocal}
\end{align} where the loss functions $\mathcal{L}_Q$ and $\mathcal{L}_D$ are defined in \eqref{eq:FedRDQNQLoss} and \eqref{eq:FRLDetectorLoss}, respectively. 
Additionally, the parameters for the target network are updated as:
\begin{equation} \label{eq:FedRDQNTargetLocal}
    \theta'_k \leftarrow \eta\theta_k + (1-\eta)\theta'_k,
\end{equation} where $\eta$ denotes a target network update rate.

\vspace{0.1cm}
\noindent\textbf{Global Update:} The server updates the global parameters by averaging the aggregated local parameters every $E$ time steps, where $E$ represents the global update interval. At each time step $t \in \Ic_{E}$, the global update is performed as:
\begin{align}\label{eq:FedRDQNGlobal}
    \bar{\theta} \leftarrow \frac{1}{K} \sum_{k=1}^{K} \theta_k,\; \bar{\theta'} \leftarrow \frac{1}{K} \sum_{k=1}^{K} \theta'_k,\; \bar{\psi} \leftarrow \frac{1}{K} \sum_{k=1}^{K} \psi_k. 
\end{align}  After global aggregation, the server broadcasts the global parameters to all $K$ agents, as defined by:
\begin{equation}\label{eq:FedRDQNBroadCast}
    (\theta_k, \theta'_k, \psi_k) \leftarrow (\bar{\theta}, \bar{\theta'}, \bar{\psi}), \quad \forall k \in [K].
\end{equation} The detailed procedure for FedRDQN is presented in Algorithm~\ref{alg:FedRDQNAlgorithm}.

\begin{algorithm}[h!]
    \caption{FedRDDPG} \label{alg:FedRDDPGAlgorithm}
    \begin{algorithmic}[1] 
        \STATE {\bf Input:} The discount factor $\gamma \in (0, 1)$, learning rates $\lambda_V, \lambda_\pi > 0$, target update rate $\eta \in (0, 1)$, mini-batch size $B$, random noise $\epsilon \sim \mathcal{N}(0, \sigma^2)$ for exploration, global update interval $E$, total time steps $T$, update starting time step $T_u$, robustness level $\omega \in [0, 1]$, and expectile level $\tau = 0.01$.

        \STATE {\bf Initialization:} For all agents $k \in [K]$, initialize network parameters $\theta_k, \psi_k, \phi_k$, set target network parameters as $\theta'_k \leftarrow \theta_k$, $\phi'_k \leftarrow \phi_k$, initialize the replay buffer $\mathcal{D}_k = \emptyset$, and set the initial state as $s_0^k \in \mathcal{S}$.

        \FOR{$t=0$ to $T-1$}
            \FOR{each $k \in [K]$}
                \STATE  $a^k_t = \pi(\cdot \mid s^k_t ; \phi_k) +  \epsilon$
                \STATE $s^k_{t+1} \sim P_k(\cdot \mid s^k_t, a^k_t)$
                \STATE $\mathcal{D}_k \leftarrow \mathcal{D}_k \cup \{ (s^k_t ,a^k_t, r(s^k_t,a^k_t), s^k_{t+1} \}$
                
                \IF{$t \ge T_u$}
                    \STATE Sample $(s, a, r, s') \sim \mathcal{D}_k$
                    \STATE Update the local parameters $\theta_k$, $\psi_k$, and $\phi_k$ via \eqref{eq:FedRDDPGQLocal}, \eqref{eq:FedRDDPGExpectileLocal}, and \eqref{eq:FedRDDPGPolicyLocal}, respectively
                    \STATE Update the target parameters $\theta'_k$ and $\phi'_k$ via \eqref{eq:FedRDDPGTargetLocal}

                \ENDIF
                \IF{The environment is truncated or done}
                    \STATE Reset the environment
                \ENDIF
            \ENDFOR
            \IF{$t + 1 \in \mathcal{I}_E$}
                \STATE Update the global parameters $(\bar{\theta}, \bar{\theta}', \bar{\psi}, \bar{\phi}, \bar{\phi}')$ via \eqref{eq:FedRDDPGGlobal}
                \STATE Broadcast the updated global parameters $(\bar{\theta}, \bar{\theta}', \bar{\psi}, \bar{\phi}, \bar{\phi}')$ to all $K$ agents
                \STATE Update the local parameters $(\theta_k, \theta'_k, \psi_k, \phi_k, \phi'_k) \leftarrow (\bar{\theta}, \bar{\theta'}, \bar{\psi}, \bar{\phi}, \bar{\phi'})$ for each agent $k \in [K]$
            \ENDIF
        \ENDFOR
    \end{algorithmic}
\end{algorithm}
\begin{figure*}[t]
    \centering
    \begin{subfigure}[b]{0.245\textwidth}
         \centering
         \includegraphics[width=\textwidth]{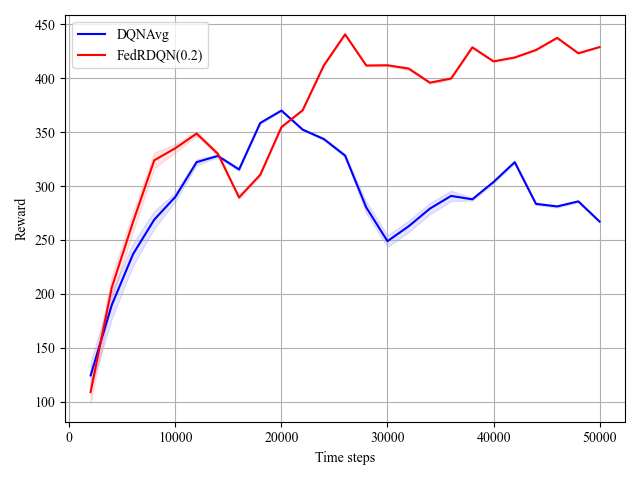}
         \caption{CartPole-v1: Length}
    \end{subfigure}
    \hfill
    \begin{subfigure}[b]{0.245\textwidth}
         \centering
         \includegraphics[width=\textwidth]{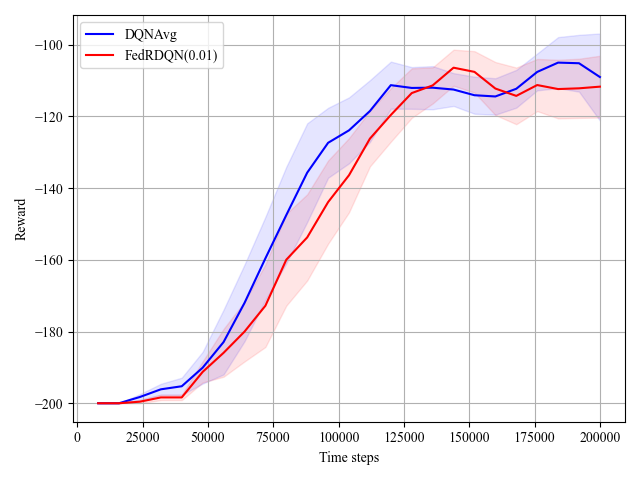}
         \caption{MountainCar-v0: Force}
    \end{subfigure}
    \hfill
    \begin{subfigure}[b]{0.245\textwidth}
         \centering
         \includegraphics[width=\textwidth]{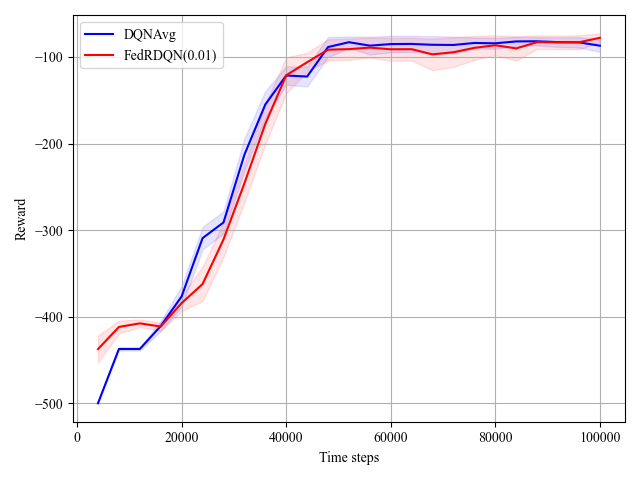}
         \caption{Acrobot-v1: Gravity}
    \end{subfigure}
    \hfill
    \begin{subfigure}[b]{0.245\textwidth}
         \centering
         \includegraphics[width=\textwidth]{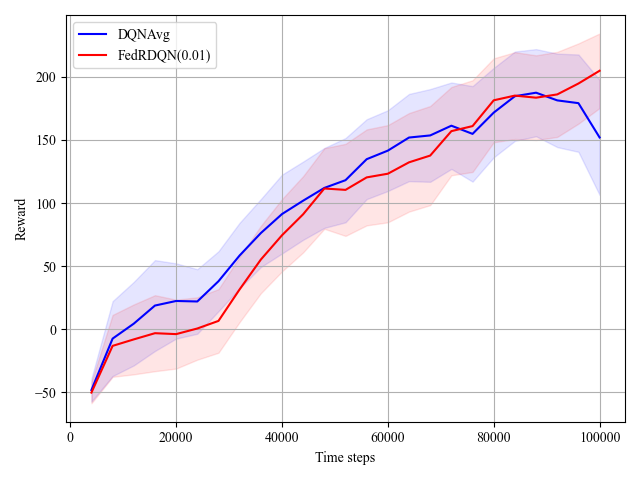}
         \caption{LunarLander-v3: Gravity}
    \end{subfigure}
    \begin{subfigure}[b]{0.245\textwidth}
         \centering
         \includegraphics[width=\textwidth]{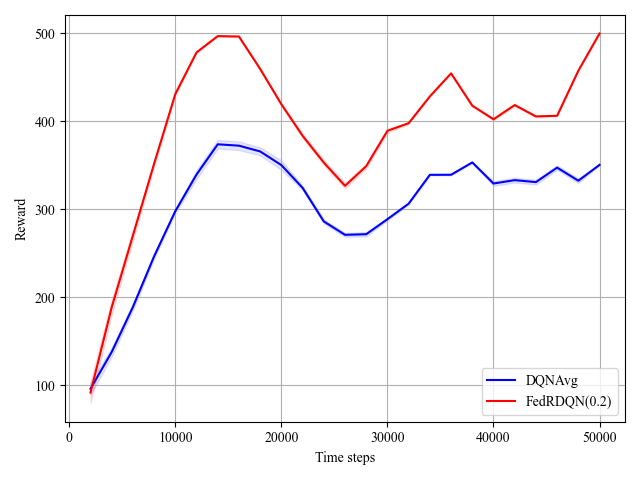}
         \caption{CartPole-v1: Mass Cart}
    \end{subfigure}
    \hfill
    \begin{subfigure}[b]{0.245\textwidth}
         \centering
         \includegraphics[width=\textwidth]{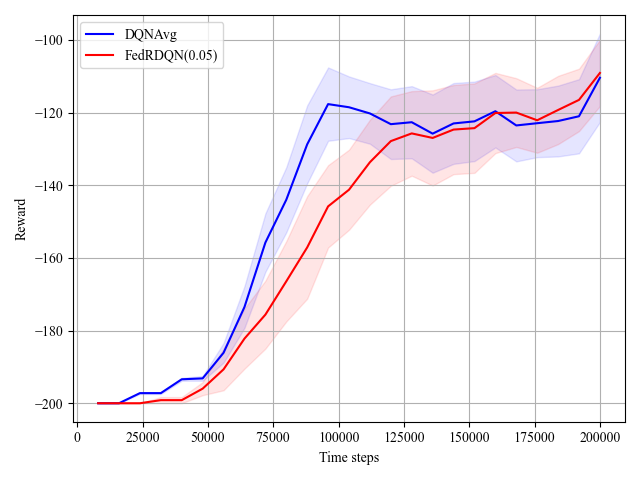}
         \caption{MountainCar-v0: Gravity}
    \end{subfigure}
    \hfill
    \begin{subfigure}[b]{0.245\textwidth}
         \centering
         \includegraphics[width=\textwidth]{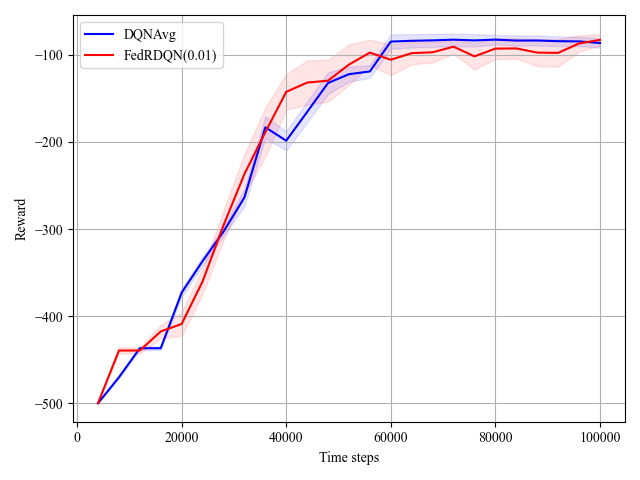}
         \caption{Acrobot-v1: Length}
    \end{subfigure}
    \hfill
    \begin{subfigure}[b]{0.245\textwidth}
         \centering
         \includegraphics[width=\textwidth]{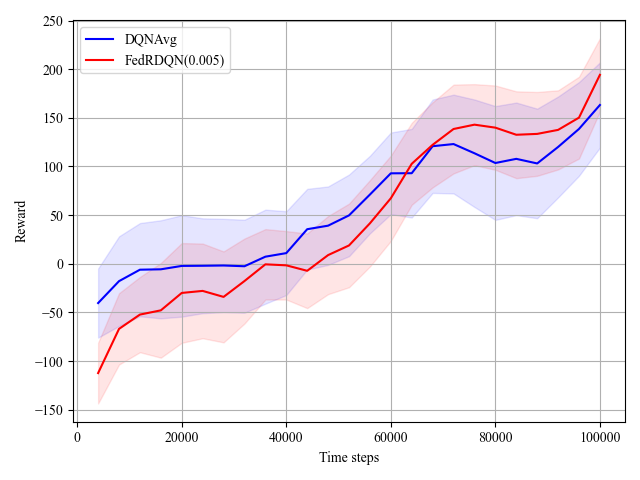}
         \caption{LunarLander-v3: Wind}
    \end{subfigure}
    \caption{{\bf Learning curves} for DQNAvg and FedRDQN evaluated on the nominal environment. The perturbed model parameter for each local environment is indicated adjacent to the environment name. The value in parentheses following each algorithm name denotes the corresponding robustness level.}
    \label{fig:dqn_training}
\end{figure*}

\begin{figure*}[t]
    \centering
    \begin{subfigure}[b]{0.245\textwidth}
         \centering
         \includegraphics[width=\textwidth]{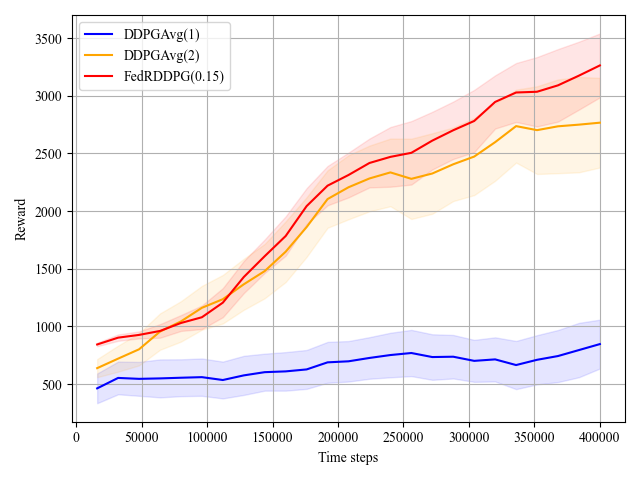}
         \caption{Ant-v4: Back Left Leg}
    \end{subfigure}
    \hfill
    \begin{subfigure}[b]{0.245\textwidth}
         \centering
         \includegraphics[width=\textwidth]{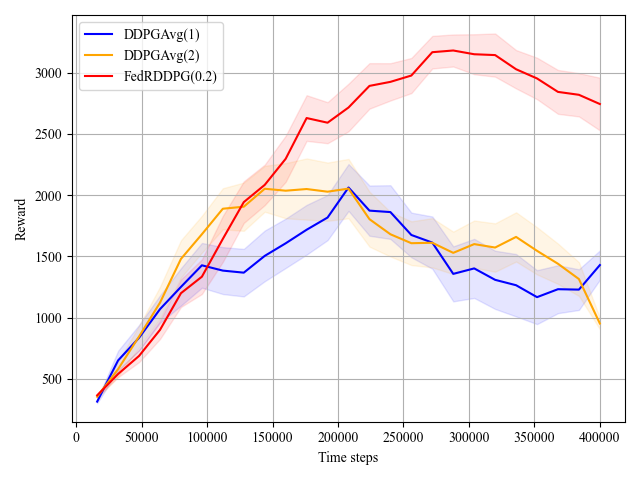}
         \caption{Hopper-v4: Thigh}
    \end{subfigure}
    \hfill
    \begin{subfigure}[b]{0.245\textwidth}
         \centering
         \includegraphics[width=\textwidth]{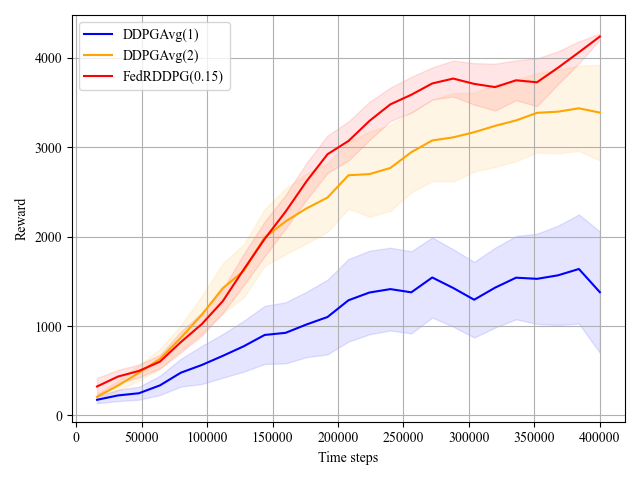}
         \caption{Walker2d-v4: Left Leg}
    \end{subfigure}
    \hfill
    \begin{subfigure}[b]{0.245\textwidth}
         \centering
         \includegraphics[width=\textwidth]{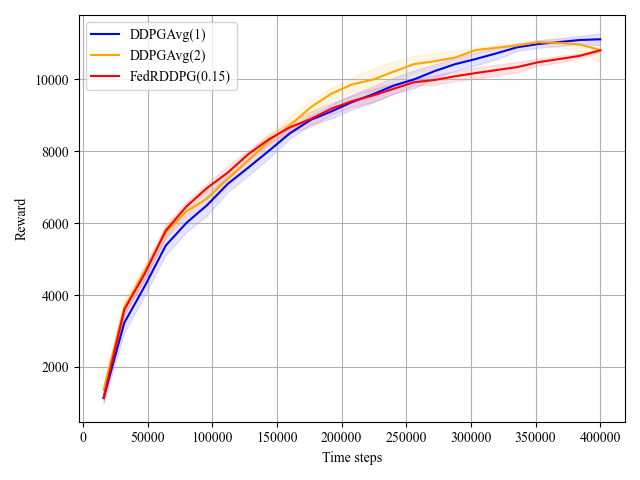}
         \caption{HalfCheetah-v4: Front Foot}
    \end{subfigure}
    \begin{subfigure}[b]{0.245\textwidth}
         \centering
         \includegraphics[width=\textwidth]{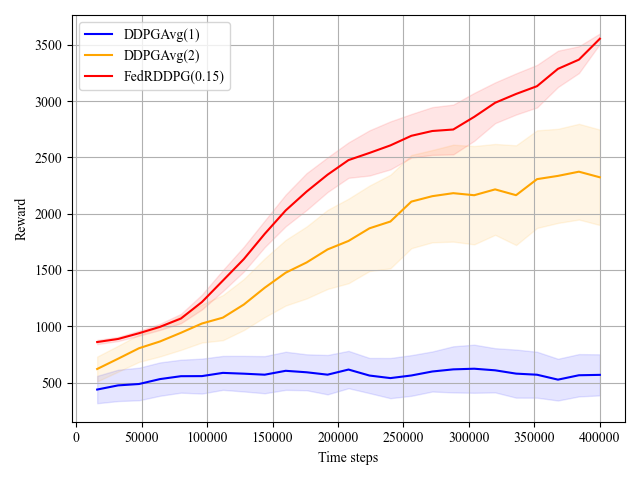}
         \caption{Ant-v4: Front Right Leg}
    \end{subfigure}
    \hfill
    \begin{subfigure}[b]{0.245\textwidth}
         \centering
         \includegraphics[width=\textwidth]{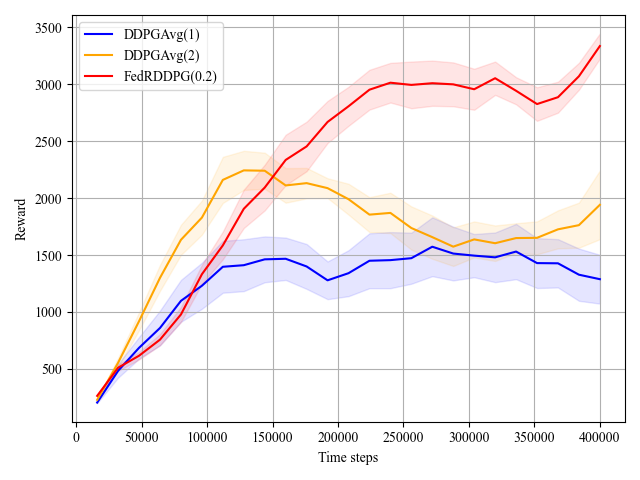}
         \caption{Hopper-v4: Torso}
    \end{subfigure}
    \hfill
    \begin{subfigure}[b]{0.245\textwidth}
         \centering
         \includegraphics[width=\textwidth]{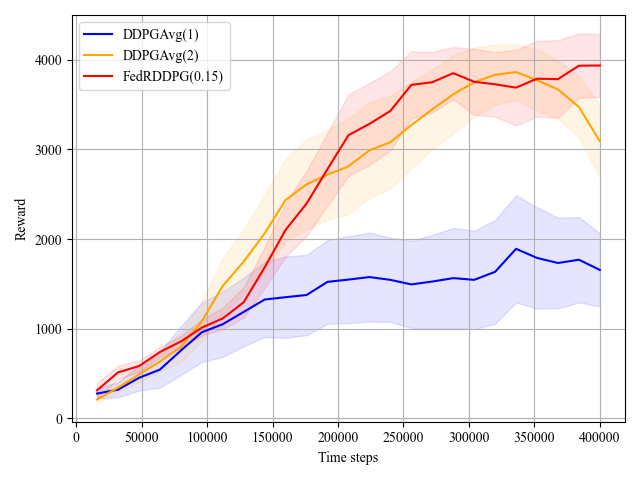}
         \caption{Walker2d-v4: Left Foot}
    \end{subfigure}
    \hfill
    \begin{subfigure}[b]{0.245\textwidth}
         \centering
         \includegraphics[width=\textwidth]{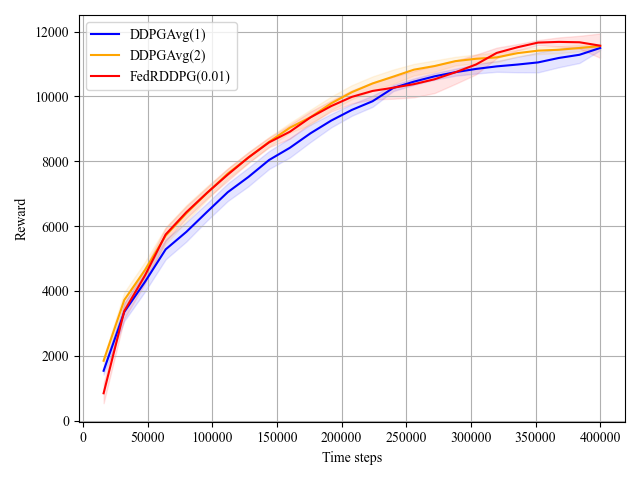}
         \caption{HalfCheetah-v4: Front Shin}
    \end{subfigure}
    \caption{{\bf Learning curves} for DDPGAvg and FedRDDPG evaluated on the nominal environment. The perturbed model parameter for each local environment is indicated adjacent to the environment name. The value in parentheses following FedRDDPG denotes the corresponding robustness level.}
    \label{fig:ddpg_training}
\end{figure*}

\subsection{FedRDDPG}\label{subsec:FedRDDPG}

DDPG is a widely recognized model-free RL algorithm specifically designed for continuous action spaces. In DDPG, both the policy function and the Q function are approximated using DNNs, allowing for applicability across various environments. Additionally, DDPG employs target networks for both the policy network and Q network to stabilize the learning process.

To address continuous control tasks, our proposed FedRQ framework can be naturally extended by integrating the DDPG algorithm. This extension, referred to as FedRDDPG, employs the policy network, denoted as $\pi(\cdot;\phi_k)$, in addition to the Q network $Q(\cdot,\cdot;\theta_k)$ and the expectile network $D(\cdot;\psi_k)$ as defined in FedRDQN. Based on this, the loss function for training the Q network of environment $k$ is given by:
\begin{equation}
    \Lc_{Q}(\theta_k)  = \mathbb{E}_{(s, a, r, s')\sim \mathcal{D}_k}\Bigg[ \Big( Q(s, a; \theta_k) - y \Big)^2 \Bigg],\label{eq:FedRDDPGQLoss}
\end{equation} where
\begin{equation}
    y = r + \gamma (1-\omega)  Q \left(s', a'; \theta'_k \right) + \gamma\omega D(s ; \psi_k),
\end{equation} and $a'= \pi(s'; \phi'_k)$. Herein,  $\theta'_k$ denote the parameters of the target Q network and $\phi'_k$ denotes the parameters of the target policy network. Additionally, the loss function for the policy network of environment $k$ is defined as:
\begin{align}\label{eq:FedRDDPGPolicyLoss}
    \mathcal{L}(\phi_k ) = \mathbb{E}_{s \sim \mathcal{D}_k} \Big[ -Q (s, \pi(s;\phi_k);\theta_k)\Big].
\end{align} 

This formulation allows each agent to train its local policy while accounting for potential variations in the environment, leveraging the expressive power of continuous control policies throughout the process. Leveraging the loss functions in \eqref{eq:FRLDetectorLoss}, \eqref{eq:FedRDDPGQLoss}, and \eqref{eq:FedRDDPGPolicyLoss}, we formally
define the local and global update processes in FedRDDPG.

\vspace{0.1cm}
\noindent{\bf Local Update:}  
Each agent $k \in [K]$ updates its local parameters at each time step $t$. Specifically, the parameters of the Q network and the expectile network are updated with a learning rate $\lambda_v$ as:
\begin{align} 
    \theta_k &\leftarrow \theta_k - \lambda_{V} \nabla \mathcal{L}_Q(\theta_k)\label{eq:FedRDDPGQLocal}\\
     \psi_k &\leftarrow \psi_k - \lambda_{V} \nabla \mathcal{L}_D(\psi_k),\label{eq:FedRDDPGExpectileLocal}
\end{align} where the loss functions $\mathcal{L}_Q$ and $\mathcal{L}_D$ are defined in \eqref{eq:FedRDDPGQLoss} and \eqref{eq:FRLDetectorLoss}, respectively.
Additionally, the parameters for the policy network are updated with a learning rate $\lambda_\pi$ as:
\begin{align} \label{eq:FedRDDPGPolicyLocal}
    \phi_k &\leftarrow \phi_k - \lambda_{\pi} \nabla \mathcal{L}_\pi(\phi_k),
\end{align}
where the loss function $\mathcal{L}_\pi$ is defined in \eqref{eq:FedRDDPGPolicyLoss}. Finally, the parameters for the target networks are updated as:
\begin{align} \label{eq:FedRDDPGTargetLocal}
    \theta'_k &\leftarrow \eta \theta_k + (1-\eta) \theta'_k \nonumber \\
    \phi'_k &\leftarrow \eta \phi_k + (1-\eta) \phi'_k,
\end{align} where $\eta$ represents a target network update rate.

\vspace{0.1cm}
\noindent{\bf Global Update:}  
The server updates the global parameters by averaging the aggregated local parameters every $E$ time steps, where $E$ represents the global update interval.  At each time step $t \in \Ic_{E}$, the global update is performed as:
\begin{align}\label{eq:FedRDDPGGlobal}
     \bar{\theta} &\leftarrow \frac{1}{K} \sum_{k=1}^{K} \theta_k,\;  \bar{\theta'} \leftarrow \frac{1}{K} \sum_{k=1}^{K} \theta'_k,\; \bar{\psi}  \leftarrow \frac{1}{K} \sum_{k=1}^{K} \psi_k\nonumber\\
     \bar{\phi} &\leftarrow \frac{1}{K} \sum_{k=1}^{K} \phi_k,\; \bar{\phi'} \leftarrow \frac{1}{K} \sum_{k=1}^{K} \phi'_k.
\end{align} After global aggregation, the server broadcasts the global parameters to all $K$ agents, as defined by:
\begin{align}
    (\theta_k, \theta'_k, \psi_k, \phi_k, \phi'_k) \leftarrow (\bar{\theta}, \bar{\theta'}, \bar{\psi}, \bar{\phi}, \bar{\phi'}), \quad \forall k \in [K].
\end{align} The complete procedure for FedRDDPG is summarized in Algorithm~\ref{alg:FedRDDPGAlgorithm}.

\begin{table*}[t]
    \centering
    \caption{{\bf Evaluation results} of DQN-based methods (DQNAvg vs. FedRDQN) on local environments. 
        \textit{Average} denotes the mean reward across all local environments, while \textit{Minimum} refers to the lowest reward observed among them corresponding to the worst-case agent.
    }
    \label{table:local_environment}
    \renewcommand{\arraystretch}{1.2}
    \begin{tabular}{ c | c | c | c | c | c} 
        \toprule
        \multirow{2}{*}{Environment} & \multirow{2}{*}{Parameter} & \multicolumn{2}{c}{{\bf DQNAvg}} & \multicolumn{2}{c}{{\bf FedRDQN}} \\
        & & Average & Minimum & Average & Minimum \\
        \midrule
        \multirow{2}{*}{Acrobot-v1} 
            & gravity & -90.2 $\pm$ 28.7 & -110.9 $\pm$ 47.5 & -84.7 $\pm$ 26.7 & -95.2 $\pm$ 30.7 \\
            & length & -83.9 $\pm$ 21.8 & -96.1 $\pm$ 26.5 & -82.9 $\pm$ 22.0 & -92.8 $\pm$ 23.8 \\
        \midrule
        \multirow{2}{*}{CartPole-v1}  
            & length & 266.9 $\pm$ 1.8 & 266.5 $\pm$ 1.7 & 428.8 $\pm$ 1.0 & 428.3 $\pm$ 0.7 \\
            & mass cart & 340.8 $\pm$ 12.3 & 309.4 $\pm$ 33.4 & 500.0 $\pm$ 0.0 & 500.0 $\pm$ 0.0 \\
        \midrule
        \multirow{2}{*}{LunarLander-v3} 
            & gravity & 152.9 $\pm$ 119.1 & 142.9 $\pm$ 100.8 & 175.0 $\pm$ 85.0 & 160.2 $\pm$ 91.2 \\
            & wind & 141.0 $\pm$ 112.1 & 134.2 $\pm$ 118.3 & 156.1 $\pm$ 108.9 & 148.7 $\pm$ 113.1 \\
        \midrule
        \multirow{2}{*}{MountainCar-v0} 
            & gravity & -135.3 $\pm$ 20.3 & -148.9 $\pm$ 16.0 & -118.6 $\pm$ 13.7 & -133.9 $\pm$ 18.4 \\
            & force & -136.7 $\pm$ 20.6 & -161.7 $\pm$ 11.2 & -129.8 $\pm$ 11.8 & -166.4 $\pm$ 10.6 \\
        \bottomrule
    \end{tabular}
\end{table*}

\begin{table*}[t]
    \centering
    \caption{{\bf Evaluation results} of DDPG-based methods (DDPGAvg(2) vs. FedRDDPG) on local environments. 
        \textit{Average} denotes the mean reward across all local environments, while \textit{Minimum} refers to the lowest reward observed among them corresponding to the worst-case agent.
    }
    \label{table:local_environment_ddpg}
    \renewcommand{\arraystretch}{1.2}
    \begin{tabular}{ c | c | c | c | c | c} 
        \toprule
        \multirow{2}{*}{Environment} & \multirow{2}{*}{Parameter} & \multicolumn{2}{c}{{\bf DDPGAvg(2)}} & \multicolumn{2}{c}{{\bf FedRDDPG}} \\
        & & Average & Minimum & Average & Minimum \\
        \midrule
        \multirow{2}{*}{Ant-v4} 
            & back left leg & 2316.9 $\pm$ 925.1 & 1569.2 $\pm$ 823.1 & 2718.0 $\pm$ 768.2 & 2415.5 $\pm$ 720.5 \\
            & front right leg & 1986.1 $\pm$ 1070.9 & 1640.7 $\pm$ 1050.1 & 2877.3 $\pm$ 676.2 & 2421.8 $\pm$ 608.4 \\
        \midrule
        \multirow{2}{*}{Hopper-v4} 
            & torso & 1946.3 $\pm$ 407.8 & 1387.3 $\pm$ 341.2 & 2604.5 $\pm$ 382.2 & 2102.5 $\pm$ 486.9 \\
            & thigh & 1037.5 $\pm$ 233.3 & 937.5 $\pm$ 103.6 & 2725.4 $\pm$ 481.1 & 2086.2 $\pm$ 698.2 \\
        \midrule
        \multirow{2}{*}{HalfCheetah-v4} 
            & front foot & 9219.2 $\pm$ 665.3 & 8352.6 $\pm$ 766.5 & 9942.9 $\pm$ 717.6 & 9499.9 $\pm$ 973.8 \\
            & front shin & 11422.8 $\pm$ 649.7 & 11295.8 $\pm$ 832.0 & 11617.8 $\pm$ 1000.3 & 11518.0 $\pm$ 1178.0 \\
        \midrule
        \multirow{2}{*}{Walker2d-v4}  
            & left leg & 3411.4 $\pm$ 1343.9 & 3186.1 $\pm$ 1482.0 & 3971.3 $\pm$ 432.6 & 3543.4 $\pm$ 376.6 \\
            & left foot & 3223.1 $\pm$ 1055.6 & 3067.7 $\pm$ 1123.9 & 4279.0 $\pm$ 744.4 & 4210.2 $\pm$ 724.0 \\
        \bottomrule
    \end{tabular}
\end{table*}

\begin{figure*}[t]
    \centering
    \begin{subfigure}[b]{0.245\textwidth}
         \centering
         \includegraphics[width=\textwidth]{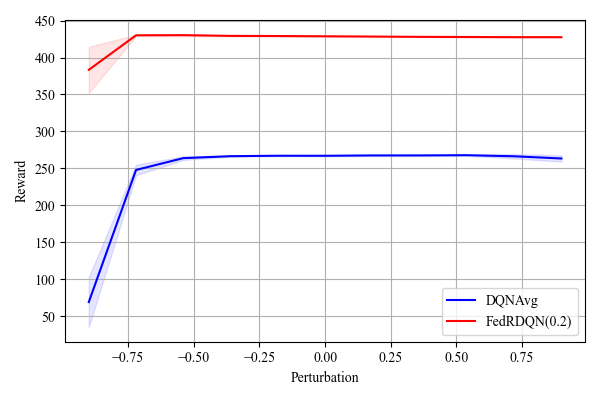}
         \caption{CartPole-v1: Length}
    \end{subfigure}
    \hfill
    \begin{subfigure}[b]{0.245\textwidth}
         \centering
         \includegraphics[width=\textwidth]{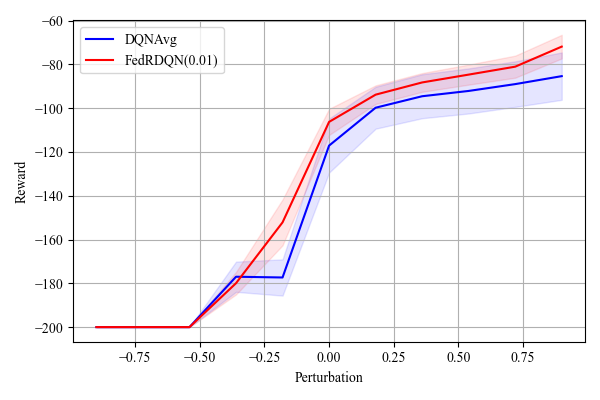}
         \caption{MountainCar-v0: Force}
    \end{subfigure}
    \hfill
    \begin{subfigure}[b]{0.245\textwidth}
         \centering
         \includegraphics[width=\textwidth]{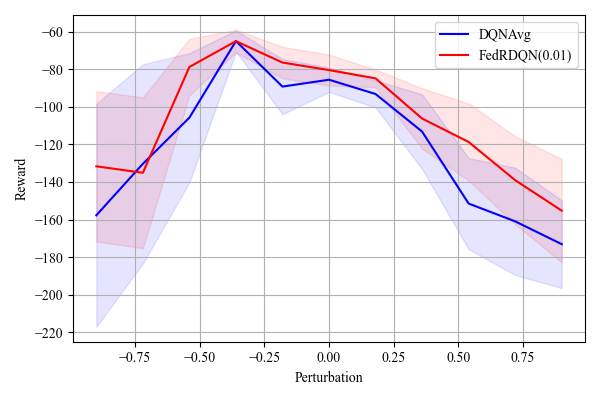}
         \caption{Acrobot-v1: Gravity}
    \end{subfigure}
    \hfill
    \begin{subfigure}[b]{0.245\textwidth}
         \centering
         \includegraphics[width=\textwidth]{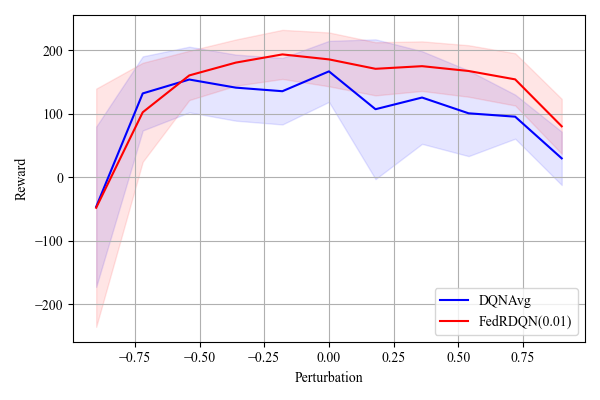}
         \caption{LunarLander-v3: Gravity}
    \end{subfigure}
    \begin{subfigure}[b]{0.245\textwidth}
         \centering
         \includegraphics[width=\textwidth]{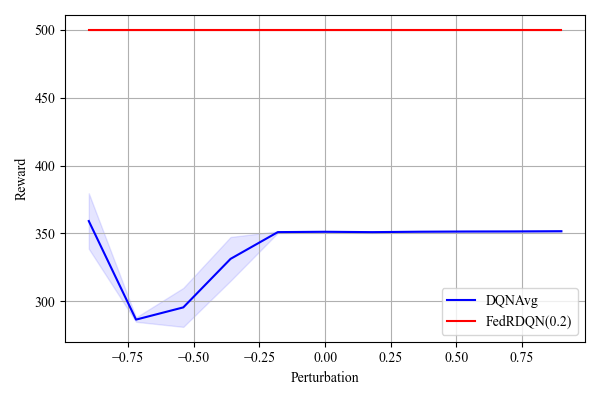}
         \caption{CartPole-v1: Mass Cart}
    \end{subfigure}
    \hfill
    \begin{subfigure}[b]{0.245\textwidth}
         \centering
         \includegraphics[width=\textwidth]{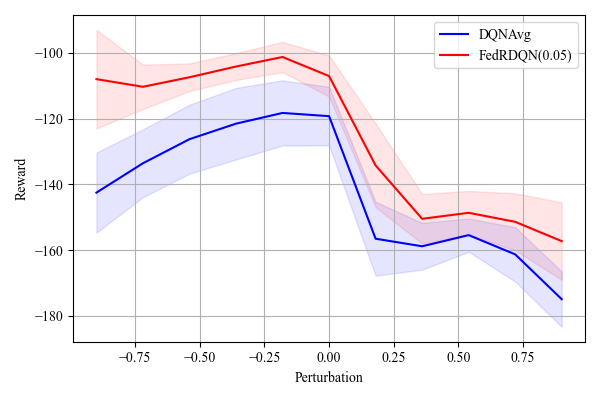}
         \caption{MountainCar-v0: Gravity}
    \end{subfigure}
    \hfill
    \begin{subfigure}[b]{0.245\textwidth}
         \centering
         \includegraphics[width=\textwidth]{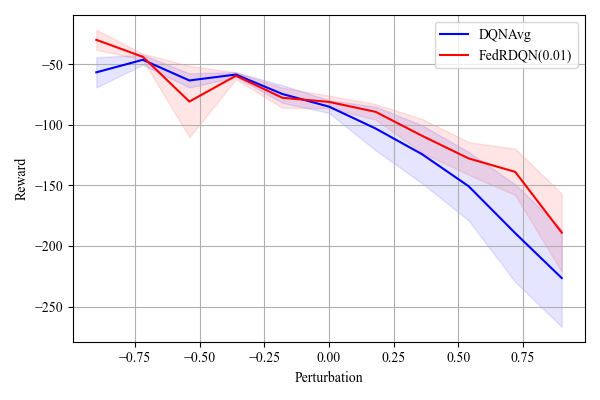}
         \caption{Acrobot-v1: Length}
    \end{subfigure}
    \hfill
    \begin{subfigure}[b]{0.245\textwidth}
         \centering
         \includegraphics[width=\textwidth]{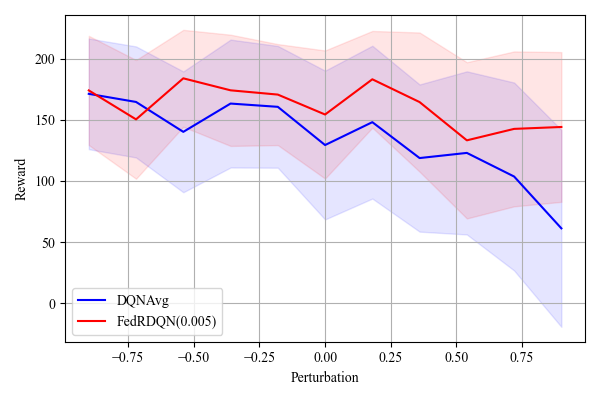}
         \caption{LunarLander-v3: Wind}
    \end{subfigure}
    \caption{{\bf Evaluation curves} for DQNAvg and FedRDQN on perturbed environments. The x-axis represents the relative deviation of the model parameter from that of the nominal environment, with zero indicating the nominal configuration. 
    The specific model parameter perturbed in each environment is annotated next to the environment name, and the value in parentheses following FedRDQN denotes its associated robustness level.}
    \label{fig:dqn_robust}
\end{figure*}

\begin{figure*}[t]
    \centering
    \begin{subfigure}[b]{0.245\textwidth}
         \centering
         \includegraphics[width=\textwidth]{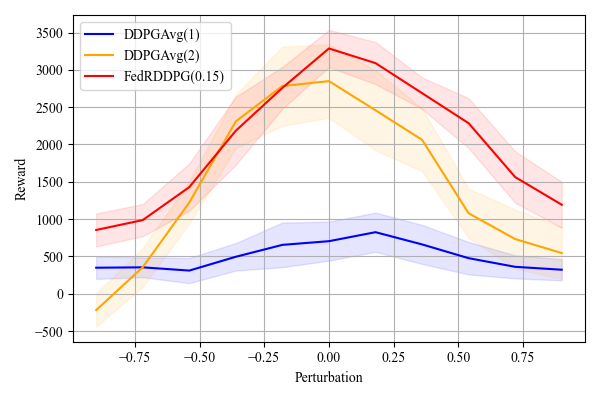}
         \caption{Ant-v4: Back Left Leg}
    \end{subfigure}
    \hfill
    \begin{subfigure}[b]{0.245\textwidth}
         \centering
         \includegraphics[width=\textwidth]{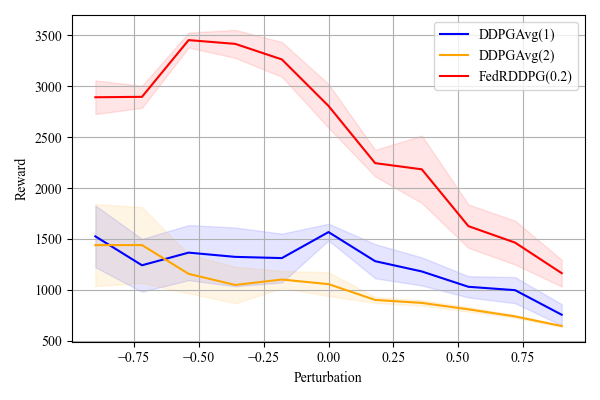}
         \caption{Hopper-v4: Thigh}
    \end{subfigure}
    \hfill
    \begin{subfigure}[b]{0.245\textwidth}
         \centering
         \includegraphics[width=\textwidth]{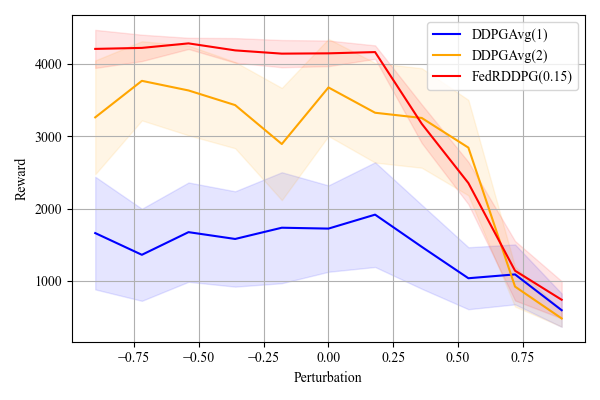}
         \caption{Walker2d-v4: Left Leg}
    \end{subfigure}
    \hfill
    \begin{subfigure}[b]{0.245\textwidth}
         \centering
         \includegraphics[width=\textwidth]{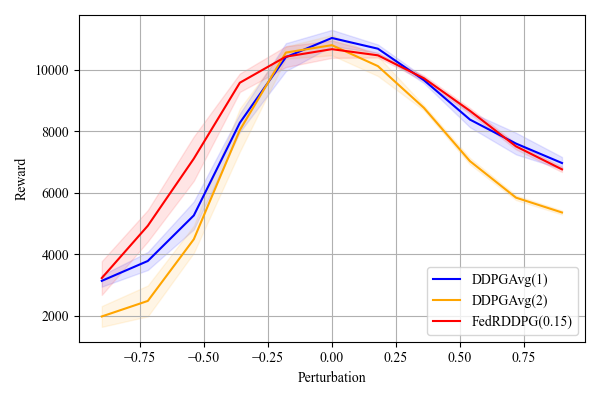}
         \caption{HalfCheetah-v4: Front Foot}
    \end{subfigure}
    \begin{subfigure}[b]{0.245\textwidth}
         \centering
         \includegraphics[width=\textwidth]{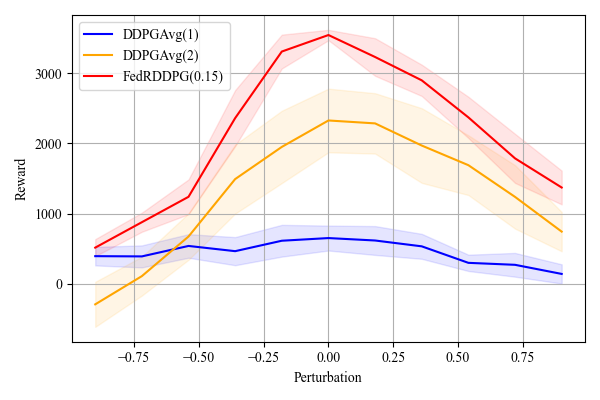}
         \caption{Ant-v4: Front Right Leg}
    \end{subfigure}
    \hfill
    \begin{subfigure}[b]{0.245\textwidth}
         \centering
         \includegraphics[width=\textwidth]{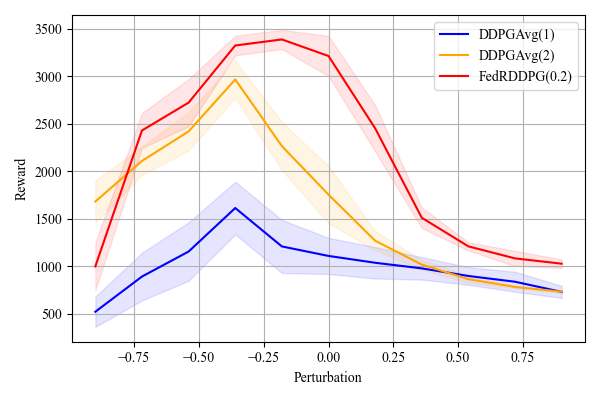}
         \caption{Hopper-v4: Torso}
    \end{subfigure}
    \hfill
    \begin{subfigure}[b]{0.245\textwidth}
         \centering
         \includegraphics[width=\textwidth]{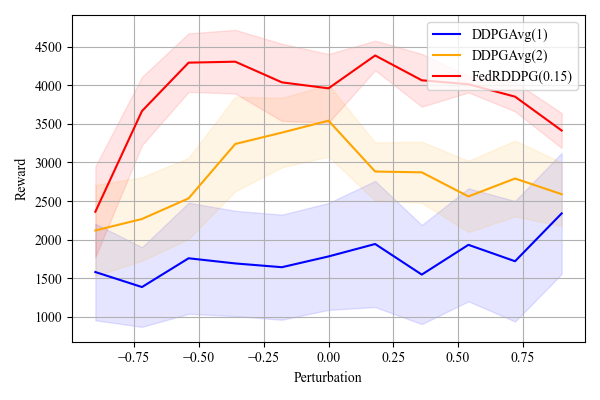}
         \caption{Walker2d-v4: Left Foot}
    \end{subfigure}
    \hfill
    \begin{subfigure}[b]{0.245\textwidth}
         \centering
         \includegraphics[width=\textwidth]{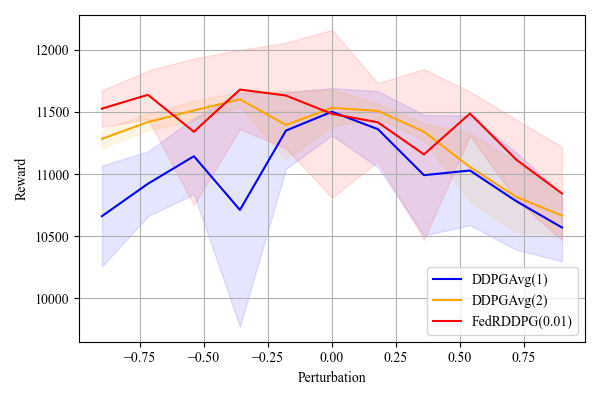}
         \caption{HalfCheetah-v4: Front Shin}
    \end{subfigure}
    \caption{{\bf Evaluation curves} for DDPGAvg and FedRDDPG on perturbed environments. The x-axis represents the relative deviation of the model parameter from that of the nominal environment, with zero indicating the nominal configuration. 
    The specific model parameter perturbed in each environment is annotated next to the environment name, and the value in parentheses following FedRDDPG denotes its associated robustness level.}
    \label{fig:ddpg_robust}
\end{figure*}

\section{Experiments} \label{sec:experiments}

We validate the effectiveness of our proposed FRL algorithms in achieving stable performance across heterogeneous environments. Experiments were conducted using a suite of environments from OpenAI Gym \cite{brockman2016openai, towers2024gymnasium}, which serve as widely accepted benchmarks in RL. These environments have also been employed in the most relevant prior work~\cite{jin2022federated}, facilitating a fair and direct comparison. For implementation, we utilize Stable-Baselines3~\cite{stable-baselines3} as our baseline framework and adopt the hyperparameter settings from RL-Zoo3~\cite{rl-zoo3}, with modifications applied only to parameters directly related to the FL setup.

We compare our proposed algorithms, FedRDQN and FedRDDPG, against the state-of-the-art FRL-EH baselines introduced in~\cite{jin2022federated}, specifically DQNAvg and DDPGAvg, respectively. However, it is well known that DDPG---the foundation algorithm underlying DDPGAvg---suffers from overestimation bias, which can substantially impair its performance~\cite{fujimoto2018addressing, kuznetsov2020controlling}. To establish a more competitive benchmark, we also consider a variant of DDPGAvg that incorporates clipped double Q-learning~\cite{fujimoto2018addressing}, which employs two Q-functions and uses the minimum of the two as the target Q-value. We refer to the original implementation as {\bf DDPGAvg(1)}, and the clipped double Q-learning variant as {\bf DDPGAvg(2)}. Our proposed FedRDDPG is evaluated against both DDPGAvg(1) and DDPGAvg(2).

To construct a federated simulation setting, we generate $K$ statistically independent environments, each of which is assigned to a corresponding local agent for interaction. To more accurately emulate real-world scenarios, we introduce heterogeneity across local environments by perturbing one of the model-defining parameters during the training phase. Specifically, each local environment is assigned to a unique, small perturbations to this parameter, resulting in distinct environment instances. Let $m$ denote a reference model parameter (e.g., the length of the pole or the mass of the cart in the CartPole-v1 environment) intended to vary across environments, such that each local environment $k$ possesses a distinct value $m_k$ for $k \in [K]$. To control the degree of variation, we define a parameter $p \in [0, 1)$, referred to as the maximum perturbation rate, which bounds the extent to which each local parameter can deviate from the reference value $m$. For each environment $k \in [K]$, we sample $n_k$ uniformly at random from the interval $(-p,p)$, and define the perturbed model parameter as $m_k = m (1 + n_k)$.  In our experiments, the hyperparameters are configured as follows: the global update interval, the number of local environments, and the maximum perturbation rate are set to:
\begin{equation}
    E=10^2,\; K=5,\; \mbox{and}\; p=0.5.
\end{equation} For environments with discrete action spaces, the mini-batch size and replay buffer size are set to $16$ and $10^3$, respectively. For continuous action spaces, these parameters are set to $64$ and $10^5$, respectively. 

\subsection{Learning Performance}

Training is conducted using five distinct random seeds to ensure the stability of the simulation results. To evaluate the performance of the global model during training, we assess its reward in the {\em nominal} environment, which is configured with the reference model parameter $m$. The resulting learning curves for DQNAvg and FedRDQN are depicted in Fig.~\ref{fig:dqn_training}, while those for DDPGAvg and FedRDDPG are illustrated in Fig.~\ref{fig:ddpg_training}.  In each figure, the solid lines represent the average rewards, and the shaded regions indicate half the standard deviation across five evaluation episodes.  For enhanced visual clarity, all curves are smoothed using a moving average filter. As demonstrated in Fig.~\ref{fig:dqn_training} and Fig.~\ref{fig:ddpg_training}, our proposed algorithms exhibit stable and consistent learning performance across all tasks, surpassing their respective baselines.

\subsection{Evaluation Performances}

To evaluate the robustness of the learned global policies within the FRL-EH framework, we assess their performance across heterogeneous local environments. Two performance metrics are considered: {\em average reward}, defined as the mean reward across all local environments, and {\em minimum reward}, defined as the lowest reward obtained among them (i.e., performance in the worst-case environment). The corresponding results are summarized in Tables~\ref{table:local_environment} and~\ref{table:local_environment_ddpg}. As shown, our proposed algorithms consistently outperform the corresponding benchmark algorithms in terms of both average and minimum performance metrics. Notably, the improvement in the minimum reward highlights the enhanced robustness of our proposed algorithms under environment heterogeneity, aligning closely with the goals of the {\em robust} FRL-EH framework. These findings demonstrate that the robust local update mechanism effectively addresses the challenges posed by heterogeneous local environments.

As previously discussed, the objective of FRL-EH is to learn a global policy that maintains robust performance across heterogeneous local environments and their plausible perturbations. To assess this robustness,  we conduct an additional evaluation of the learned policy by deploying it in a series of testing environments, where the model parameter is systematically varied from $0.1m$ to $1.9m$. The corresponding evaluation results for DQNAvg and FedRDQN are shown in Fig.~\ref{fig:dqn_robust}, while those for DDPGAvg and FedRDDPG are illustrated in Fig.~\ref{fig:ddpg_robust}. As with the training curves, the solid lines indicate the average rewards, and the shaded areas represent half the standard deviation across evaluation episodes. All curves are smoothed using a moving average filter to enhance visual clarity. These results demonstrate that the proposed FedRDQN and FedRDDPG consistently exhibit improved robustness compared to their respective benchmarks, DQNAvg and DDPGAvg, under various model perturbations. These findings confirm the effectiveness of the proposed robust local update mechanism in preserving policy performance under environmental heterogeneity.

\section{Conclusions} \label{sec:conclusion}

We introduced the {\em robust} FRL-EH framework, which is based on a novel global objective function that effectively captures both statistical heterogeneity across local environments and practically plausible perturbations through a covering set. Within this framework, we proposed a new tabular algorithm, termed FedRQ, and provided a theoretical proof of its asymptotic convergence to an optimal solution. Unlike existing methods, FedRQ adopts a robust local update mechanism that explicitly accounts for variability across environments, thereby improving the stability and robustness of FRL. We further extended FedRQ to function approximations by integrating its core principles with widely used deep RL algorithms, and empirically demonstrated its effectiveness across a variety of challenging tasks characterized by environment heterogeneity. Collectively, these findings underscore the practical viability and improved performance of the proposed approach. A promising direction of future research is to investigate more efficient representations of the covering set, with the goal of further enhancing learning performance in increasingly diverse and complex real-world scenarios.

\bibliographystyle{IEEEtran}
\bibliography{CS_REF}

\end{document}